\documentclass[10pt,twocolumn,letterpaper]{article}

\usepackage[pagenumbers]{wacv} 

\usepackage{graphicx}
\usepackage{amsmath}
\usepackage{amssymb}
\usepackage{booktabs}

\usepackage{marvosym}
\usepackage{siunitx}
\usepackage{array,multirow}
\usepackage{colortbl}
\usepackage{xcolor}
\usepackage{arydshln}
\usepackage{caption}
\usepackage[export]{adjustbox}
\usepackage{subcaption}
\usepackage{algorithm}
\usepackage{algpseudocode}
\mathchardef\mhyphen="2D
\usepackage[colorinlistoftodos,prependcaption,textsize=tiny]{todonotes}

\usepackage[pagebackref,breaklinks,colorlinks]{hyperref}

\usepackage[capitalize]{cleveref}
\crefname{section}{Sec.}{Secs.}
\Crefname{section}{Section}{Sections}
\Crefname{table}{Table}{Tables}
\crefname{table}{Tab.}{Tabs.}

\def\wacvPaperID{1115} 
\def\confName{WACV}
\def\confYear{2024}

\begin{document}

\title{Efficient Multi-task Uncertainties for Joint Semantic Segmentation \\ and Monocular Depth Estimation}

\author{Steven Landgraf \hspace{1em} Markus Hillemann \hspace{1em} Theodor Kapler \hspace{1em} Markus Ulrich\\
Institute of Photogrammetry and Remote Sensing (IPF)\\
Karlsruhe Institute of Technology\\
{\tt\small (steven.landgraf, markus.hillemann, markus.ulrich)@kit.edu}\\
{\tt\small theodor.kapler@student.kit.edu}
}
\maketitle

\begin{abstract}
Quantifying the predictive uncertainty emerged as a possible solution to common challenges like overconfidence or lack of explainability and robustness of deep neural networks, albeit one that is often computationally expensive. Many real-world applications are multi-modal in nature and hence benefit from multi-task learning. In autonomous driving, for example, the joint solution of semantic segmentation and monocular depth estimation has proven to be valuable. In this work, we first combine different uncertainty quantification methods with joint semantic segmentation and monocular depth estimation and evaluate how they perform in comparison to each other. Additionally, we reveal the benefits of multi-task learning with regard to the uncertainty quality compared to solving both tasks separately. Based on these insights, we introduce EMUFormer, a novel student-teacher distillation approach for joint semantic segmentation and monocular depth estimation as well as efficient multi-task uncertainty quantification. By implicitly leveraging the predictive uncertainties
of the teacher, EMUFormer achieves new state-of-the-art results on Cityscapes and NYUv2 and additionally estimates high-quality predictive uncertainties for both tasks that are comparable or superior to a Deep Ensemble despite being an order of magnitude more efficient.
\end{abstract}

\section{Introduction}
Because of their unparalleled performance in fundamental perception tasks like semantic segmentation \cite{minaee2020ImageSegmentation} or monocular depth estimation \cite{dong2022towards}, deep neural networks are increasingly being deployed in real-time and safety-critical applications like autonomous driving \cite{mcallister2017ConcreteProblems}, industrial inspection \cite{steger2018MachineVision, heizmann2022implementing}, and automation \cite{landgraf2023segmentation}. However, they often suffer from overconfidence \cite{guo2017CalibrationModerna}, lack explainability \cite{gawlikowski2022SurveyUncertainty}, and struggle to distinguish between in-domain and out-of-domain samples \cite{lee2018TrainingConfidencecalibrated}, which is of paramount importance for applications where prediction reliability is crucial. Since incorrect predictions can lead to severe consequences, previous work suggests that quantifying the uncertainty inherent to a model's prediction is a promising endeavour to make such applications safer \cite{landgraf2023dudes, leibig2017LeveragingUncertainty, lee2018TrainingConfidencecalibrated, mukhoti2018evaluating, mukhoti2023deep, landgraf2023u, loquercio2020general}. In autonomous driving, for instance, the car could provide feedback to the driver when it is uncertain or preemptively make risk-averse predictions based on the uncertainty.

In recent years, a number of promising uncertainty quantification methods have been proposed to make deep neural networks more robust \cite{mackay1992PracticalBayesian, gal2016DropoutBayesian, lakshminarayanan2017SimpleScalable, valdenegro2023sub, van2020uncertainty, liu2020simple, mukhoti2023deep, amini2020deep}. Unfortunately, these methods either introduce technical complexity or require computationally expensive sampling from a stochastic process to estimate the uncertainty of a prediction. Additionally, they do not consider that real-world applications, like robotics \cite{nekrasov2019real} or autonomous driving \cite{LiangfuDriving}, are multi-modal in nature and benefit from multi-task learning, especially within the context of semantic segmentation and monocular depth estimation \cite{nekrasov2019real,LiangfuDriving}. Although there have been successful attempts at making uncertainty quantification methods more efficient through the concept of knowledge distillation \cite{landgraf2023dudes, besnier2021learning, Holder_2021_ICCV, Shen_2021_WACV, simpson2022learning}, they have thereby either focused on semantic segmentation \cite{landgraf2023dudes, besnier2021learning, Holder_2021_ICCV,Shen_2021_WACV} or monocular depth estimation \cite{Shen_2021_WACV, simpson2022learning}. This represents a notable research gap in the current literature. 

\begin{table}[t!]
\begin{center}
\begin{adjustbox}{width=\linewidth}
\setlength\extrarowheight{1mm}
\begin{tabular}{l|cc|cc|cc|c}
& Seg. & Pred. Unc. & Depth & Pred. Unc. & Parameters & FLOPs & FPS \\ \hline
a) SegFormer-B2\cite{xie2021segformer} & $\checkmark$ & $\times$ & $\times$ & $\times$ & 27.3M & 72.6G & 55.3 \\
b) DepthFormer-B2 & $\times$ & $\checkmark$ & $\times$ & $\times$ & 27.3M & 72.1G & 57.1 \\
c) SegDepthFormer-B2 & $\checkmark$ & $\times$ & $\checkmark$ & $\times$ & 30.5M & 120.1G & 44.8 \\ \cdashline{1-8}
DE of a) & $\checkmark$ & $\checkmark$ & $\times$ & $\times$ & 273.6M & 726.4G & 5.6 \\
DE of b) & $\times$ & $\times$ & $\checkmark$ & $\checkmark$ & 273.5M & 720.8G & 7.2 \\
DE of c) & $\checkmark$ & $\checkmark$ & $\checkmark$ & $\checkmark$ & 305.1M & 1201.1G & 4.9 \\ \cdashline{1-8}
EMUFormer-B2 (Ours) & $\checkmark$ & $\checkmark$ & $\checkmark$ & $\checkmark$ & 30.5M & 120.1G & 44.8 \\
\end{tabular}
\end{adjustbox}
\end{center}
\caption{Overview of the segmentation (Seg.), depth estimation (Depth) and uncertainty quantification (Pred. Unc.) capabilities as well as the respective number of parameters, FLOPs and FPS for different single-task and multi-task models and their respective Deep Ensemble (DE) versions with 10 members. SegFormer \cite{xie2021segformer} and DepthFormer represent single-task models, whereas SegDepthFormer and EMUFormer depict multi-task models. B2 represents the medium-sized encoder of SegFormer, which was used for all models. Results are based on single-scale inference conducted on the NYUv2 \cite{silberman2012indoor} dataset using an NVIDIA A100 GPU.}
\label{table: comparison}
\end{table}

In this work, we conduct a comprehensive series of experiments to study multi-task uncertainties and propose a novel student-teacher distillation approach for joint semantic segmentation and monocular depth estimation as well as efficient multi-task uncertainty quantification. Our contributions can summarized as follows:
\begin{itemize}
    \itemsep0em 
    \item We propose a novel student-teacher distillation approach for \textbf{E}fficient \textbf{M}ulti-task \textbf{U}ncertainties for joint semantic segmentation and monocular depth estimation with a modern Vision-Trans\textbf{former}, which we call \textbf{EMUFormer}.
    \item We show that by implicitly leveraging the predictive uncertainties during training, EMUFormer can achieve new state-of-the-art results on Cityscapes and NYUv2.
    \item We combine different uncertainty quantification methods with joint semantic segmentation and monocular depth estimation and evaluate how they perform in comparison to each other.
    \item We reveal the benefits of multi-task learning with regard to the uncertainty quality compared to solving semantic segmentation and monocular depth estimation separately.
\end{itemize}
As Table \ref{table: comparison} demonstrates, EMUFormer estimates high-quality predictive uncertainties for both tasks that are comparable to the Deep Ensemble teacher despite being an order of magnitude more efficient. 


\section{Related Work}\label{sec: related work}
In this section, we summarize the related work on joint semantic segmentation and monocular depth estimation, uncertainty quantification, and knowledge distillation. 

\subsection{Joint Semantic Segmentation and Monocular Depth Estimation}
Semantic segmentation and monocular depth estimation are both fundamental problems in image understanding that involve pixel-wise predictions based on a single input image. Motivated by the strong correlation and complementary properties of the two tasks, multiple previous works have focused on solving both tasks in a joint manner \cite{wang2015towards,mousavian2016joint,jiao2018look,xu2018pad,liu2018collaborative,lin2019depth,nekrasov2019real,he2021sosd,gao2022ci,ji2023semantic,kendall2018multi,liu2019end,bruggemann2021exploring,xu2022mtformer,bruggemann2020automated}. To limit the scope of this literature review, we refrain from covering other multi-task approaches with joint representation sharing \cite{zhang2021survey} or methods that leverage the depth map to improve the semantic segmentation prediction \cite{hu2018rgb, wang2021brief}.

In their pioneering work, Wang et al. \cite{wang2015towards} propose a unified framework for semantic segmentation and monocular depth prediction through joint training and applying a two-layer hierarchical conditional random field to enforce synergy between global and local predictions. Similarly, Liu et al. \cite{liu2018collaborative} use a conditional random field that fuses the feature maps from both tasks. In contrast, Mousavian et al. \cite{mousavian2016joint} train parts of the model for each task separately and then fine-tune the full model on both tasks with a single loss function. On a similar note, Xu et al. \cite{xu2018pad} propose a multi-task prediction-and-distillation network, which first predicts a set of intermediate auxiliary tasks. These intermediate outputs are then utilized as multi-modal input for the final task - a concept also followed by Vandenhende et al. \cite{vandenhende2020mti}. The idea of knowledge distillation is also used by Nekrasov et al. \cite{nekrasov2019real}, primarily focusing on real-time estimation without specifically delving into uncertainty quantification. Jiao et al. \cite{jiao2018look} introduce an attention-driven loss that does not treat all pixels in an image equally to mutually improve semantic segmentation and monocular depth estimation. In a similar way, Bruggemann et al. \cite{bruggemann2021exploring} and Liu et al. \cite{liu2019end} build on the idea of introducing attention mechanisms into the architecture to improve results. Comparably, Gao et al. \cite{gao2022ci} propose a shared attention block with contextual supervision next to a feature-sharing module and a consistency loss. In a follow-up work, they extend their approach by incorporating confidences into their losses to improve the performance \cite{gao2022predictive}. Similarly, Kendall et al. \cite{kendall2018multi} utilize the homoscedastic uncertainty, which they define as a task-dependent uncertainty that captures the relative confidence between tasks, to weight the individual losses. Finally, there are multiple works \cite{lin2019depth,he2021sosd,ji2023semantic} that propose specialized architectures, where they either improve the feature extraction by separating the relevant features for one task from the features which are relevant for both tasks \cite{lin2019depth} or exploit geometric constraints by integrating the information of the objectness \cite{he2021sosd} or apply a randomly-weighted training strategy to balance the losses and gradients impartially and dynamically \cite{ji2023semantic}. 

Remarkably, most of the discussed approaches use out-of-date architectures and require complex adaptions to either the model, the training process, or both. In order to push the state-of-the-art forward, we adapt a modern Vision-Transformer-based architecture similar to Xu et al. \cite{xu2022mtformer}. In order to maintain methodological simplicity and transparency of the results, we refrain from introducing cross-task attention mechanisms, contrastive self-supervised learning algorithms, and the loss weighting strategy of \cite{kendall2018multi}, and nevertheless achieve superior results. However, these strategies could also be applied to our method, potentially further improving the results.

\subsection{Uncertainty Quantification}\label{sec: uq}
A large variety of uncertainty quantification methods \cite{mackay1992PracticalBayesian, gal2016DropoutBayesian, lakshminarayanan2017SimpleScalable, valdenegro2023sub, van2020uncertainty, liu2020simple, mukhoti2023deep, amini2020deep} have been developed to compensate for the above-mentioned shortcomings of deep neural networks. The predictive uncertainty can be decomposed into aleatoric and epistemic uncertainty \cite{gal2016uncertainty}. Aleatoric uncertainty captures the irreducible data uncertainty, which, for example, can be introduced by image noise or noisy labels as a result of imprecise measurements. Epistemic uncertainty accounts for the model uncertainty, which can be reduced by using more or better training data \cite{gal2016uncertainty,kendall2017CVUncertainties}. Disentangling these two uncertainty components can be essential for applications such as active learning \cite{gal2017deep} or the detection of out-of-distribution samples \cite{schwaiger2020uncertainty}. For instance, active learning benefits from avoiding inputs with high aleatoric uncertainty unless they exhibit high epistemic uncertainty, which is vital for model improvement \cite{gal2017deep,kendall2017CVUncertainties}. 

Most well-known uncertainty quantification methods require multiple forward passes at test time, making them computationally expensive. For instance, Gal and Ghahramani \cite{gal2016DropoutBayesian} propose Monte Carlo Dropout (MCD) as an approximation of a stochastic Gaussian process. While dropout is usually only used for regularization during training \cite{srivastava2014Dropout}, MCD applies this technique during test time to sample from the posterior distribution of the predictions at test time. Although MCD is easy to implement and thus very popular, Deep Ensembles \cite{lakshminarayanan2017SimpleScalable} are commonly regarded as the state-of-the-art approach for uncertainty quantification across varying tasks \cite{ovadia2019DatasetShift, wursthorn2022, gustafsson2020evaluating}. They consist of an ensemble of trained models that generate diverse predictions due to the introduction of randomness through random weight initialization or different data augmentations during training \cite{fort2020DeepEnsembles}.

Multiple forward passes at test time render the aforementioned methods impracticable or even unusable for real-time applications because of their high computational cost. Consequently, there has been an increased interest in deterministic single forward-pass methods that demand less overhead. For example, Van Amersfoort et al. \cite{van2020uncertainty} and Liu et al. \cite{liu2020simple} consider distance-aware output layers for quantifying the predictive uncertainty. Even though these methods provide a computationally more efficient approach, they are not competitive with the current state-of-the-art and require significant modifications to the training process \cite{mukhoti2023deep}. By using Gaussian Discriminant Analysis post-training for feature-space density estimation, Mukhoti et al. \cite{mukhoti2023deep} simplify the aforementioned approaches. Although they manage to perform on par with a Deep Ensemble in some settings, their method requires performing Gaussian Discriminant Analysis after training, which adds complexity. In contrast, Valdenegro-Toro \cite{valdenegro2023sub} proposes a simple, yet effective approximation to Deep Ensembles, where the ensemble covers only a subset of layers instead of the whole model. These so-called Deep Sub-Ensembles (DSE) enable a trade-off between uncertainty quality and computational cost \cite{valdenegro2023sub}.

To the best of our knowledge, quantifying predictive uncertainties in joint semantic segmentation and monocular depth estimation has not been explored yet. To this end, we compare multiple uncertainty quantification methods for this task and investigate how multi-task learning influences the quality of uncertainty estimates in comparison to solving both tasks separately.

\subsection{Knowledge Distillation}
Knowledge distillation, introduced by Hinton et al. \cite{hinton2015DistillingKnowledgea}, involves transferring the knowledge from a complex model (teacher) to a typically smaller model (student), aiming to enhance the student's performance on a given task by imitating the predictions of the teacher \cite{hinton2015DistillingKnowledgea} or transferring knowledge from intermediate features \cite{romero2015FitNetsHints}. More recent work has adapted the concept of knowledge distillation to enable real-time uncertainty quantification. While some previous work employs MCD to estimate uncertainties for the student to learn \cite{Shen_2021_WACV,gurau2018dropout,besnier2021learning}, the majority proposes to use a Deep Ensemble \cite{Holder_2021_ICCV,Deng_2021_ICCV,landgraf2023dudes,simpson2022learning,malinin2019EnsembleDistributiona}. Among these, Deng et al. \cite{Deng_2021_ICCV} are the only ones to consider a multi-task problem by looking at emotion recognition.

To enable real-time uncertainty quantification in joint semantic segmentation and monocular depth estimation, we propose EMUFormer, a novel student-teacher distillation approach that aims to preserve both prediction and uncertainty quality without introducing a speed-penalty during inference. 

\section{Methodology}\label{sec: metholodogy}
In the following, we provide an overview of the methodology of this paper, describe the baseline models that we use to analyse the uncertainties of joint semantic segmentation and monocular depth estimation. We will also explain our student-teacher distillation approach for efficient multi-task uncertainties. 

\subsection{Overview}
This paper can broadly be categorized into two parts: First, we evaluate how multi-task learning influences the uncertainty quality. Second, we propose EMUFormer, a novel student-teacher distillation approach for efficient multi-task uncertainties.

\textbf{Multi-task Uncertainty Evaluation.} Drawing from the related work on uncertainty quantification (Section \ref{sec: uq}), we evaluate Deep Ensembles (DEs) \cite{lakshminarayanan2017SimpleScalable}, Monte Carlo Dropout (MCD) \cite{gal2016DropoutBayesian}, and Deep Sub-Ensembles (DSEs) \cite{valdenegro2023sub}. The choice is motivated by their simplicity, ease of implementation, parallelizability, minimal tuning requirements, and representation of the current state-of-the-art in uncertainty quantification. Moreover, applying these approaches to both semantic segmentation and monocular depth estimation is straightforward, which is not the case for the other aforementioned uncertainty quantification approaches \cite{mukhoti2023deep, van2020uncertainty, liu2020simple, amini2020deep}.

To explore the impact of multi-task learning on uncertainty quality, we conduct all of the evaluations using three models:

\begin{enumerate}
    \itemsep0em     
    \item \textbf{SegFormer} \cite{xie2021segformer}: An efficient semantic segmentation Vision Transformer.
    \item \textbf{DepthFormer}: An efficient monocular depth estimation model Vision Transformer.
    \item \textbf{SegDepthFormer}: A joint model addressing both semantic segmentation and monocular depth estimation.
\end{enumerate}

We derive the latter two, DepthFormer and SegDepthFormer, from the SegFormer \cite{xie2021segformer} architecture. Key modifications will be explained in Section \ref{sec: depthformer} and \ref{sec: segdepthformer} respectively.

\textbf{EMUFormer.} In order to achieve efficient multi-task uncertainties without sacrificing neither prediction performance nor uncertainty quality, we propose EMUFormer. EMUFormer applies student-teacher distillation as a two-step framework: First, we train an adequate teacher with ground truth labels that is able to quantify high-quality uncertainties. Subsequently, we train a student with the same ground truth labels while distilling the teacher's uncertainties.

\begin{figure}[t!]
    \centering
    \includegraphics[width=0.99\linewidth]{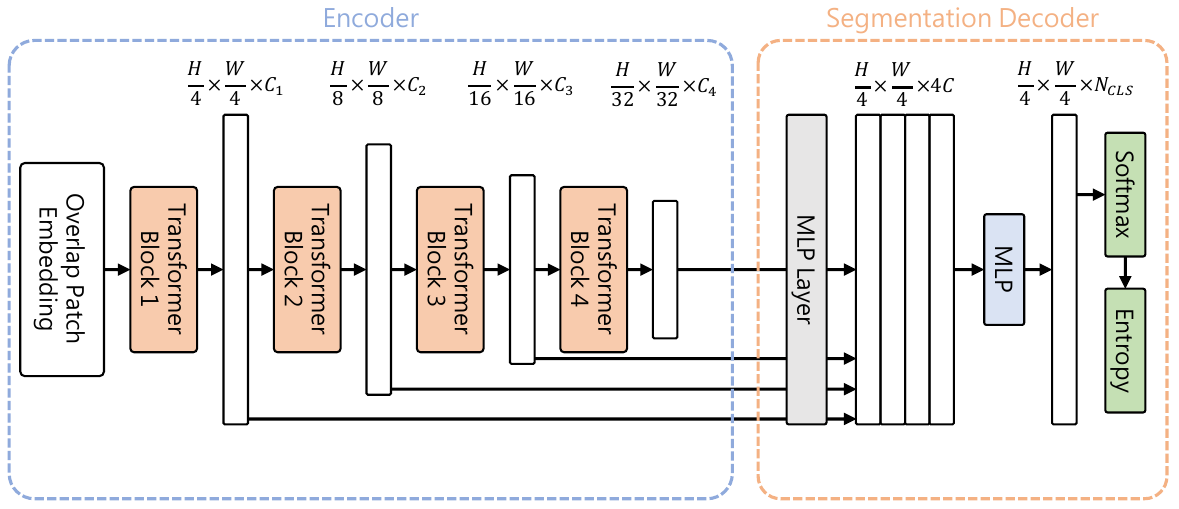}
    \caption{A schematic overview of the SegFormer \cite{xie2021segformer} architecture. The model consists of two main modules: A hierarchical Transformer-based encoder that generates high-resolution coarse features and low-resolution fine features and a lightweight all-MLP segmentation decoder.}
    \label{fig: SegFormer}
\end{figure}

\subsection{Baseline Models}
Hereinafter, we go over the three baseline models, SegFormer \cite{xie2021segformer}, DepthFormer, and SegDepthFormer. For all of the three models, we will shortly describe their architecture, illustrate the training criterion, and how we obtain a measurement for the uncertainty. While these models are capable of estimating the aleatoric uncertainty \cite{kendall2017CVUncertainties, lakshminarayanan2017SimpleScalable}, they are not able to quantify the more complete predictive uncertainty, which includes the epistemic uncertainty. For this, one of the aforementioned uncertainty quantification methods has to be used. 

\begin{figure}[t!]
    \centering
    \includegraphics[width=0.99\linewidth]{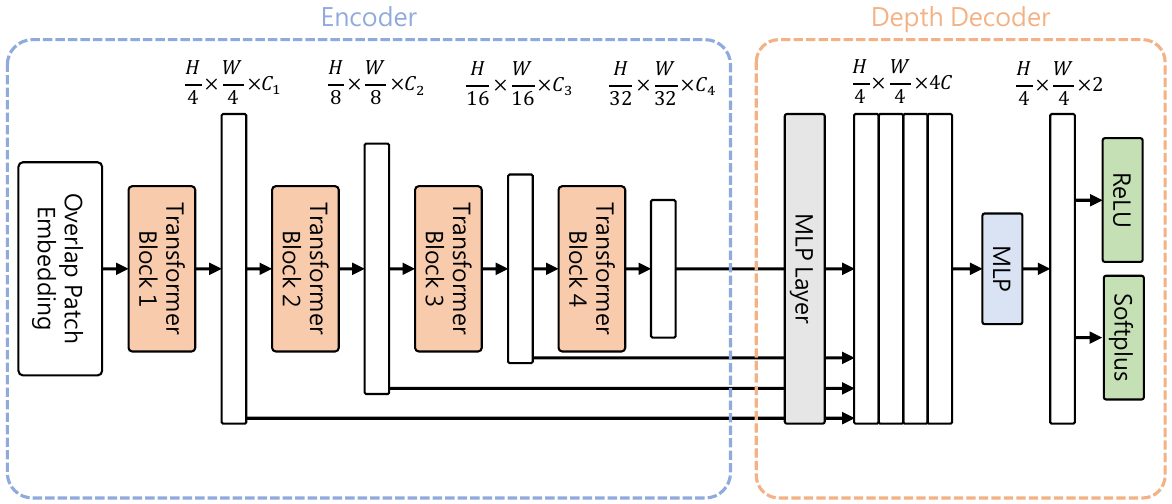}
    \caption{A schematic overview of our DepthFormer architecture. Being derived from SegFormer \cite{xie2021segformer}, it consists of two main modules: A hierarchical Transformer-based encoder that generates high-resolution coarse features and low-resolution fine features and a lightweight all-MLP depth decoder.}
    \label{fig: DepthFormer}
\end{figure}

\begin{figure*}[t!]
\centering
\includegraphics[width=0.99\linewidth]{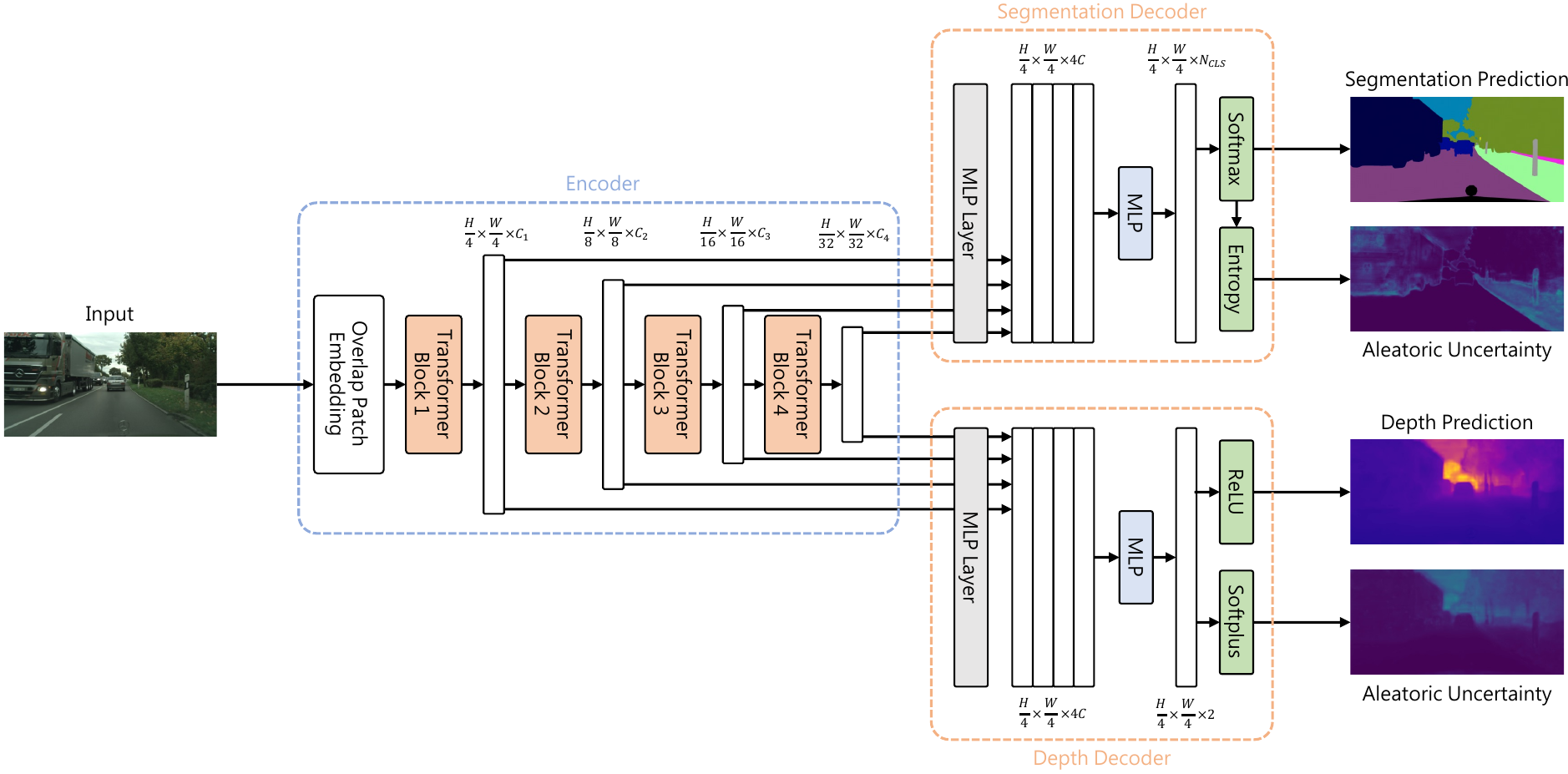}
\caption{A schematic overview of the SegDepthFormer architecture. The model combines the SegFormer \cite{xie2021segformer} architecture with a lightweight all-MLP depth decoder.}
\label{fig: SegDepthFormer}
\end{figure*}

\subsubsection{SegFormer}\label{sec: segformer}
\textbf{Architecture.} For the semantic segmentation task, we use SegFormer \cite{xie2021segformer}, a modern Transformer-based architecture that stands out because of its high efficiency and performance. Thus, it is particularly suitable for real-time uncertainty quantification. As depicted in Figure \ref{fig: SegFormer}, SegFormer consists of two main modules: A hierarchical Transformer-based encoder that generates high-resolution coarse features and low-resolution fine features and a lightweight all-MLP segmentation decoder. The latter fuses the multi-level features of the encoder to produce a final segmentation prediction with the softmax activation function, which can be formulated as:
\begin{equation}\label{eq: softmax}
    p(z) = \frac{e^{z_i}}{\sum_{k=1}^{K} e^{z_k}},  
\end{equation}
where $p(z)$ are the class probabilities of the softmax function that exponentiates each of the $K$ elements of the input vector $x$, often referred to as logits, and then normalizes the results to obtain a probability distribution. Since SegFormer \cite{xie2021segformer} only outputs logits at a $\frac{H}{4} \times \frac{W}{4}$ resolution given an input image of size $H \times W$, we use bilinear interpolation \cite{xie2021segformer} before applying the softmax function on $z$ to obtain the original resolution for the final segmentation prediction. 

\textbf{Training Criterion.} For the objective function during training, we use the well-known categorical Cross-Entropy loss
\begin{equation}\label{eq: ce}
    \mathcal{L}_\mathrm{CE} = - \frac{1}{N} \sum_{n=1}^{N} \sum_{c=1}^{C} y_{n,c} \cdot \log(p(z)_{n, c})
    \enspace,
\end{equation}
where $\mathcal{L}_\mathrm{CE}$ is the Cross-Entropy loss for a single image, $N$ is the number of pixels in the image, $C$ is the number of classes, $y_{n,c}$ is the corresponding ground truth label, and $p(z)_{n,c}$ is the predicted softmax probability.

\textbf{Aleatoric Uncertainty.} We compute the predictive entropy
\begin{equation}\label{eq: entropy}
    H(p(z)) = -\sum_{c=1}^{C} p(z)_c \cdot \log(p(z)_c)
    \enspace,
\end{equation}
which serves as the aleatoric uncertainty \cite{kendall2017CVUncertainties}. 

\subsubsection{DepthFormer}\label{sec: depthformer}
\textbf{Architecture.} Inspired by the efficiency and performance of SegFormer \cite{xie2021segformer}, we propose DepthFormer for monocular depth estimation. As Figure \ref{fig: DepthFormer} shows, we use the same hierarchical Transformer-based encoder as SegFormer to generate high-level and low-level features. Similarly, those multi-level features are fused in an all-MLP decoder. In contrast to the segmentation decoder, the depth decoder differs by having two output channels: one for the predictive mean $\mu(z)$ and one for the predictive variance $s^2(z)$ \cite{loquercio2020general}.

\textbf{Predictive Mean.} The first output channel uses a Rectified Linear Unit (ReLU) output activation function
\begin{equation}\label{eq: relu}
    \mu(z) = \max(0, z)
    \enspace,
\end{equation}
which serves as the predictive mean for monocular depth estimation.

\textbf{Predictive Variance.} The second output channel applies a Softplus activation
\begin{equation}\label{eq: variance}
    s^2(z) = \log(1 + e^z)
    \enspace,
\end{equation}
which is a smooth approximation of the ReLU function with the advantage of being differentiable, also at $z = 0$. Empirically, we found Softplus to work better than ReLU for the predictive variance, following the work by Lakshminarayanan et al. \cite{lakshminarayanan2017SimpleScalable}. 

\begin{figure*}[t!]
    \centering
    \includegraphics[width=0.99\linewidth]{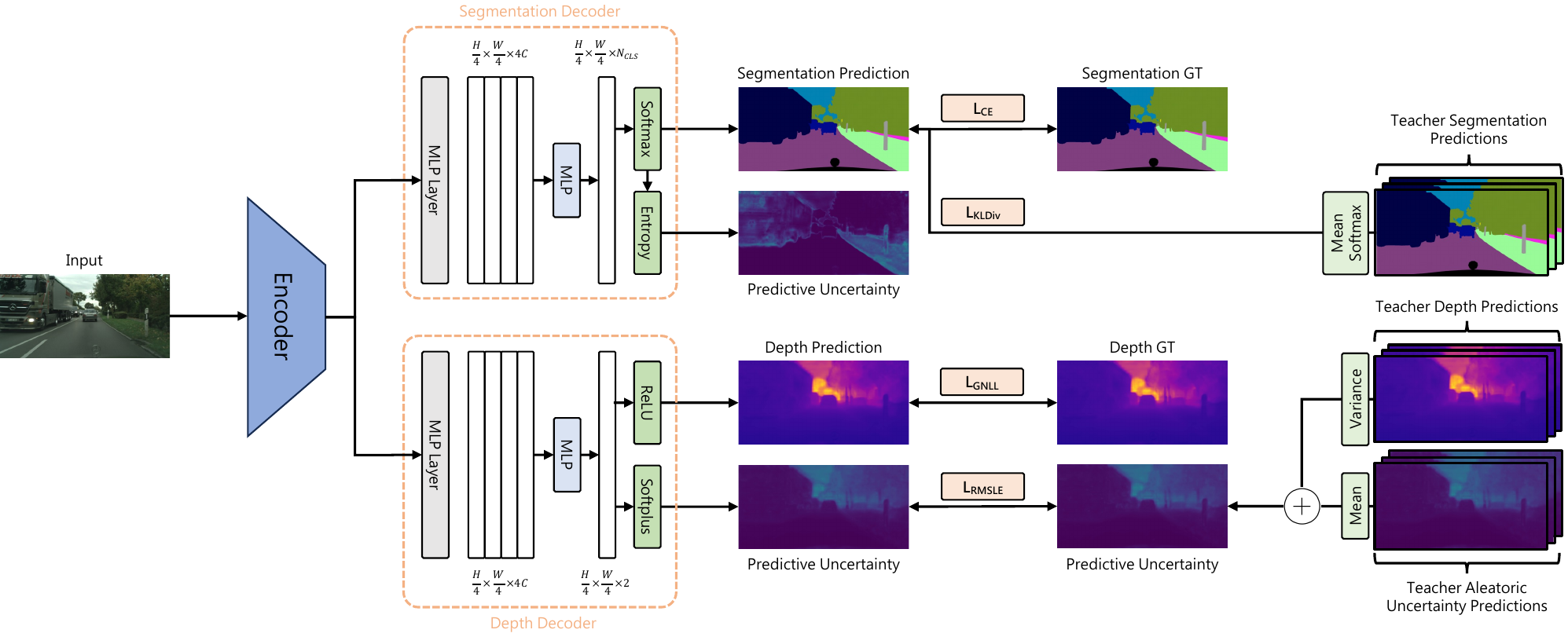}
    \caption{A schematic overview of EMUFormer. In comparison to our proposed SegDepthFormer, EMUFormer utilizes two additional losses that distill the predictive uncertainties of the teacher into the student model.}
    \label{fig: EMUFormer}
\end{figure*}

\textbf{Training Criterion.} For regression tasks, neural networks typically output only a predictive mean $\mu(z)$ and the parameters are, in the most straightforward approach, optimized by minimizing the mean squared error (MSE). However, the MSE does not cover uncertainty. Therefore, we follow the approach of Nix and Weigend \cite{nix1994estimating} instead: By treating the neural networks prediction as a sample from a Gaussian distribution with the predictive mean $\mu(z)$ and corresponding predictive variance $s^2(z)$, we can minimize the Gaussian Negative Log-Likelihood (GNLL) loss, which can be formulated as:
\begin{equation}\label{eq: gnll}
    \mathcal{L}_\mathrm{GNLL} = \frac{1}{2} \left( \frac{(y - \mu(z))^2}{s^2(z)} + \log(s^2(z)) \right)
    \enspace,
\end{equation}
where $y$ is the the ground truth depth.  

\textbf{Aleatoric Uncertainty.} Through GNLL minimization, DepthFormer does not only optimize the predictive means, but also inherently learns the corresponding variances, which can be interpreted as the aleatoric uncertainty \cite{kendall2017CVUncertainties,loquercio2020general}.

\subsubsection{SegDepthFormer.}\label{sec: segdepthformer}
\textbf{Architecture.} In order to solve semantic segmentation and monocular depth estimation in a joint manner, we propose SegDepthFormer. The architecture, which is shown in Figure~\ref{fig: SegDepthFormer}, comprises three modules: a hierarchical Transformer-based encoder, an all-MLP segmentation decoder, and an all-MLP depth decoder. The encoder and segmentation decoder are adapted from SegFormer \cite{xie2021segformer} (Section \ref{sec: segformer}), while the depth decoder is from DepthFormer (Section \ref{sec: depthformer}). Both decoders fuse the multi-level features obtained through the shared encoder to predict a final segmentation mask and a pixel-wise depth estimation, respectively.

\textbf{Training Criterion.} SegDepthFormer is trained to minimize the weighted sum of the two previously described objective functions:
\begin{equation}
    \mathcal{L} = \mathcal{L}_\mathrm{CE} + w_1 \mathcal{L}_\mathrm{GNLL}
    \enspace,
\end{equation}
where $w_1$ is a simple weighting factor. Because both loss values are of similar magnitude, we set $w_1 = 1$. However, tuning $w_1$ might slightly improve SegDepthFormer's performance. 

\textbf{Aleatoric Uncertainty.} The respective aleatoric uncertainty is obtained by computing the predictive entropy $H(p(z))$ (see Equation \ref{eq: entropy}) for the segmentation task or by the predictive variance $s^2(z)$ (see Equation \ref{eq: variance}), which is learned implicitly through the optimization of $\mathcal{L}_\mathrm{GNLL}$.

\subsection{EMUFormer}
In the following, we explain our student-teacher distillation framework for efficient multi-task uncertainties, which we call EMUFormer.
Our objective with EMUFormer is threefold: 
\begin{enumerate}
    \itemsep0em     
    \item Achieve state-of-the-art joint semantic segmentation and monocular depth estimation results.
    \item Estimate well-calibrated predictive uncertainties for both tasks.
    \item Avoid introducing additional computational overhead during inference. 
\end{enumerate}

In order to achieve these goals, EMUFormer employs a two-step student-teacher distillation framework:
\begin{enumerate}
    \itemsep0em     
    \item Training a teacher with ground truth labels.
    \item Training the student with ground truth labels while distilling the teacher's predictive uncertainties.
\end{enumerate}

\textbf{Teacher.} Although our framework is flexible with regard to the type of teacher, we use a DE that is known for producing high-quality estimates \cite{ovadia2019DatasetShift, wursthorn2022, gustafsson2020evaluating}. 

\textbf{Student.} We propose employing the SegDepthFormer architecture for the student model due to its simplicity, performance, and efficiency. In principle, though, any architecture capable of outputting a semantic segmentation mask along with a predictive mean and variance for monocular depth estimation is suitable.

\textbf{Distillation Approach.} To efficiently estimate predictive uncertainties for semantic segmentation and monocular depth estimation, EMUFormer utilizes student-teacher distillation. Figure \ref{fig: EMUFormer} shows a schematic overview of EMUFormer.  The training is performed with two additional uncertainty-related losses compared to the regular SegDepthFormer. To compute both predictive uncertainties we compute multiple prediction samples from the teacher. Additionally, we add color jittering as an additional data augmentation to the teacher's input $\Tilde{x}$. Previous work showed that this is helpful when the training dataset is used for training and distillation to prevent the student from underestimating the epistemic uncertainty of the teacher \cite{Shen_2021_WACV, landgraf2023dudes}. The color jitter causes the teacher's uncertainty distribution on the training dataset to be more closely related to the test-time distribution.

\textbf{Segmentation Uncertainty Loss.} The segmentation uncertainty knowledge of the teacher model is transferred into the student model by using the Kullback-Leibler divergence loss:
\begin{equation}
    \mathcal{L}_\mathrm{KL} = \sum_{c=1}^{C} q_c(\Tilde{z}) \cdot \log\left(\frac{q_c(\Tilde{z})}{p_c(z)}\right)
    \enspace,
\end{equation}
where $\Tilde{z}$ are the logits based on the perturbed input image, $q_c(\Tilde{z})$ is the teacher's mean softmax probability map, and $p_c(z)$ is the student's softmax probability map. Minimizing this loss ensures that the student learns to match the well-calibrated softmax probabilities provided by the teacher, allowing the predictive entropy $H(p(z))$ (see Equation \ref{eq: entropy}) to capture the underlying predictive uncertainty.


\textbf{Depth Uncertainty Loss.} Because it is not possible to match two distributions for the unbound uncertainties in the regression task, we introduce the root mean squared logarithmic error (RMSLE) for the depth uncertainty distillation:
\begin{equation}
    \resizebox{0.86\linewidth}{!}{$\mathcal{L}_\mathrm{RMSLE} = \sqrt{\frac{1}{N} \sum_{n=1}^{N} \left(\log(\sigma_n^2(\Tilde{z}) + 1) - \log(s_n^2(z) + 1)\right)^2}$}
    \enspace,
\end{equation}
where $\sigma_n^2(\Tilde{z})$ is the teacher's predictive uncertainty and $s_n^2(z)$ is the student's predictive uncertainty estimate. The natural logarithm penalizes underestimations more than overestimations, thereby providing special attention to the pixels with higher uncertainties. Minimizing the depth uncertainty loss trains the student to mimic the predictive uncertainty of the teacher. Consequently, the second output channel of the decoder does not only output the aleatoric uncertainty anymore, but rather the more meaningful predictive uncertainty, which additionally covers the epistemic uncertainty.

We follow Loquercio et al. \cite{loquercio2020general} to calculate the predictive uncertainty of the teacher with:
\begin{equation}\label{eq: pred uncertainty depth}
    \sigma^2(\Tilde{z}) = \frac{1}{T} \sum_{t=1}^{T} s_t^2(\Tilde{z}) + \frac{1}{T} \sum_{t=1}^{T} \left( \mu_t(\Tilde{z}) - \bar{\mu_t}({\Tilde{z}}) \right)^2
    \enspace,
\end{equation}
where $T$ is the number of prediction samples from the teacher, $v(\Tilde{z})$ is the predictive variance (see Equation \ref{eq: variance}, $\mu(\Tilde{z})$ is the predictive mean of a sample, and $\bar{\mu}({\Tilde{z}})$ is the mean predictive mean across all samples.

\textbf{Training Criterion.} In summary, EMUFormer is trained to minimize the weighted sum of four objective functions: 
\begin{equation}\label{eq: emuformer}
    \resizebox{0.86\linewidth}{!}{$\mathcal{L} = \mathcal{L}_\mathrm{CE} + w_1 \mathcal{L}_\mathrm{GNLL} + w_2 \mathcal{L}_\mathrm{KL} + w_3 \mathcal{L}_\mathrm{RMSLE}.$}
\end{equation}
By setting $w_1 = w_3 = 1$ and $w_2 = 10$, we obtain good results across all of our experiments. However, depending on the application, tuning these hyperparameters may further enhance performance.

\section{Experimental Setup}\label{sec: Exp Setup}
\textbf{Datasets.} We conduct all experiments on Cityscapes \cite{cordts2016CityscapesDataset} and NYUv2 \cite{silberman2012indoor}. Cityscapes, with 2975 training and 500 validation images, is a popular urban street scene benchmark dataset. Notably, the depth values are based on the disparity of stereo camera images. NYUV2 contains 795 training and 654 testing images of indoor scenes.  

\textbf{Data Augmentations.} Regardless of the trained model, we apply a very common data augmentation strategy:

\begin{enumerate}
    \itemsep0em
    \item Random scaling with a factor between $0.5$ and $2.0$.
    \item Random cropping with a crop size of $768\times768$ pixels on Cityscapes and $480 \times 640$ pixels on NYUv2.
    \item Random horizontal flipping with a flip chance of $50\%$. 
\end{enumerate}

\textbf{Implementation Details.} For all training processes, we use AdamW \cite{loshchilov2017decoupled} optimizer with a base learning rate of 0.00006 and employ a polynomial rate scheduler:
\begin{equation}
lr = lr_\mathrm{base} \cdot (1 - \frac{\mathrm{iteration}}{\mathrm{total\:iterations}})^{0.9}
\enspace,
\end{equation}
where $lr$ is the current learning rate and $lr_{base}$ is the initial base learning rate. Besides, we use a batch size of 8 and train on four NVIDIA A100 GPUs with 40 GB of memory using mixed precision \cite{micikevicius2017mixed}. The encoders of the baseline models are initialized with weights pre-trained on ImageNet \cite{deng2009ImageNetLargescale} and then trained for 250 epochs on Cityscapes and for 100 epochs on NYUv2, respectively. In contrast, EMUFormer is initialized with the weights of a pre-trained SegDepthFormer and fine-tuned for 100 epochs on both datasets. Unless otherwise noted, we use the SegFormer-B2 \cite{xie2021segformer} backbone for all experiments. We do not adopt any of the widely-used methods such as OHEM \cite{shrivastava2016training}, auxiliary losses, class imbalance compensation, or sliding window testing to keep our approach as simple and transparent as possible.

\textbf{Metrics.} For quantitative evaluations of the semantic segmentation task, we report the mean Intersection over Union (mIoU), also known as the Jaccard Index. Additionally, we use the Expected Calibration Error (ECE) \cite{naeini2015obtaining} to evaluate the calibration of the softmax probabilities. For the monocular depth estimation task, we use the common root mean squared error (RMSE). Finally, we employ the following uncertainty evaluation metrics proposed by Mukhoti and Gal \cite{mukhoti2018evaluating}:

\begin{enumerate}
    \itemsep0em     
    \item $p(accurate|certain)$: The probability that the model is accurate on its output given that the uncertainty is below a certain threshold.
    \item $p(uncertain|inaccurate)$: The probability that the uncertainty of the model exceeds a certain threshold given that the prediction is inaccurate.
    \item $PAvPU$: The combination of both cases, i.e. accurate$|$certain and inaccurate$|$uncertain. 
\end{enumerate}

Although these metrics have originally been proposed for semantic segmentation \cite{mukhoti2018evaluating}, we also use them to evaluate the depth regression uncertainties. Since one cannot simply determine whether a depth prediction is accurate, we apply the following formula: 
\begin{equation}
    \max\left(\frac{\mu(z)}{y}, \frac{y}{\mu(z)}\right) = \delta_{1} < 1.25
    \enspace,
\end{equation}
where $\mu(z)$ is the predicted depth value of a pixel and $y$ is the corresponding ground truth depth \cite{ming2021deep}. $\delta_{1}$ serves as a standard metric for quantifying the accuracy of monocular depth estimation models, using $1.25$ as the threshold to determine whether a depth prediction is accurate or not. In contrast, $\delta_{2}$ and $\delta_{3}$ are less strict, typically utilizing thresholds of $1.25^2$ and $1.25^3$, respectively.

For the sake of simplicity and to simulate real-world employment, we set the uncertainty threshold to the mean uncertainty of a given image for all evaluations.

\textbf{Monte Carlo Dropout.} MCD depends primarily on the number of dropout layers, where they are inserted inside the network, and most-importantly the dropout rate. Since the original SegFormer \cite{xie2021segformer} already applies dropout layers throughout the entire network, we follow their work and only consider two dropout rates, 20\% and 50\%. We sample ten times to obtain the prediction and predictive uncertainty \cite{gal2016DropoutBayesian, Shen_2021_WACV, gustafsson2020evaluating}.

\textbf{Deep Sub-Ensemble.} Consistent with the DEs and MCD, we train the DSE with ten decoder heads for each task on top of a shared encoder \cite{valdenegro2023sub}. During training, we only optimize a single decoder head per training batch and alternate between them. Thereby, we aim to introduce as much randomness as possible, analogous to the training of DEs. For inference, we utilize all decoder heads, of course.

\textbf{Deep Ensemble.} DEs achieve the best results if they are trained to explore diverse modes in function space, which we accomplish by randomly initializing all decoder heads, by using random augmentations, and by applying random shuffling of the training data points \cite{lakshminarayanan2017SimpleScalable, fort2020DeepEnsembles}. Unless otherwise noted, we report results of a DE with ten members, following the suggestions of previous work \cite{lakshminarayanan2017SimpleScalable, fort2020DeepEnsembles, landgraf2023dudes}.

\textbf{Predictions.} Regardless of the uncertainty quantification method, we report the results of the mean prediction. For the semantic segmentation task, we compute the mean softmax probability of all samples. For the monocular depth estimation task, we first apply ReLU (see Equation \ref{eq: relu}) and then compute the mean depth of the corresponding samples.

\textbf{Uncertainty.} For the semantic segmentation task, we compute the predictive entropy (see Equation \ref{eq: entropy}) based on the mean softmax probabilities as a measure for the predictive uncertainty \cite{mukhoti2018evaluating}. For the depth estimation task, however, we calculate the predictive uncertainty based on the mean predictive variance and the variance of the depth predictions of the samples (see Equation \ref{eq: pred uncertainty depth}) \cite{loquercio2020general}.

\section{Joint Uncertainty Evaluation}\label{sec: joint uncertainty evaluation}

\begin{table*}[t!]
\begin{center}
\begin{adjustbox}{width=\linewidth}
\setlength\extrarowheight{1mm}
\begin{tabular}{ll|ccccc|cccc|c}
& \multirow{2}{*}{} & \multicolumn{5}{c|}{Semantic Segmentation} & \multicolumn{4}{c|}{Monocular Depth Estimation} & \multirow{2}{*}{Inference Time [ms]}\\ \cdashline{3-11}
& & mIoU $\uparrow$ & ECE $\downarrow$ & p(acc/cer) $\uparrow$ & p(inacc/unc) $\uparrow$ & PAvPU $\uparrow$ & RMSE $\downarrow$ & p(acc/cer) $\uparrow$ & p(inacc/unc) $\uparrow$ & PAvPU $\uparrow$ & \\ \hline \hline
\multirow{3}{*}{\rotatebox[origin=c]{90}{Baseline}}
& \multicolumn{1}{|l|}{SegFormer} & 0.772 & 0.033 & 0.882 & 0.395 & 0.797 & - & - & - & - & 17.90 $\pm$ 0.47 \\ 
& \multicolumn{1}{|l|}{DepthFormer} & - & - & - & - & - & 7.452 & 0.749 & 0.476 & 0.766 & 17.59 $\pm$ 0.82 \\ 
& \multicolumn{1}{|l|}{SegDepthFormer} & 0.738 & 0.028 & 0.913 & 0.592 & 0.826 & 7.536 & 0.745 & 0.472 & 0.762 & 22.04 $\pm$ 0.27 \\ \cdashline{1-12}
\multirow{3}{*}{\rotatebox[origin=c]{90}{\shortstack{MCD \\ (20\%)}}}
& \multicolumn{1}{|l|}{SegFormer} & 0.759 & \textbf{0.007} & 0.883 & 0.424 & 0.780 & - & - & - & - & 177.13 $\pm$ 0.64 \\ 
& \multicolumn{1}{|l|}{DepthFormer}  & - & - & - & - & - & 7.956 & 0.749 & 0.555 & 0.739 & 139.32 $\pm$ 0.78 \\ 
& \multicolumn{1}{|l|}{SegDepthFormer}  & 0.738 & 0.020 & 0.911 & 0.592 & 0.803 & 7.370 & 0.761 & 0.523 & 0.757 & 202.23 $\pm$ 0.39 \\ \cdashline{1-12}
\multirow{3}{*}{\rotatebox[origin=c]{90}{\shortstack{MCD \\ (50\%)}}}
& \multicolumn{1}{|l|}{SegFormer}  & 0.662 & 0.028 & 0.883 & 0.485 & 0.760 & - & - & - & - & 176.98 $\pm$ 0.53 \\ 
& \multicolumn{1}{|l|}{DepthFormer}  & - & - & - & - & - & 21.602 & 0.181 & 0.366 & 0.431 & 139.81 $\pm$ 1.20 \\ 
& \multicolumn{1}{|l|}{SegDepthFormer}  & 0.640 & 0.021 & 0.906 & 0.616 & 0.782 & 8.316 & 0.733 & \textbf{0.558} & 0.723 & 203.82 $\pm$ 0.81 \\ \cdashline{1-12}
\multirow{3}{*}{\rotatebox[origin=c]{90}{\shortstack{DSE}}}
& \multicolumn{1}{|l|}{SegFormer} & 0.772 & 0.037 & 0.890 & 0.456 & 0.797 & - & - & - & - & 132.30 $\pm$ 3.16 \\ 
& \multicolumn{1}{|l|}{DepthFormer}  & - & - & - & - & - & \textbf{7.036} & 0.762 & 0.467 & 0.772 & 91.82 $\pm$ 2.01 \\ 
& \multicolumn{1}{|l|}{SegDepthFormer}  & 0.749 & 0.009 & \textbf{0.931} & \textbf{0.696} & \textbf{0.844} & 7.441 & 0.751 & 0.463 & 0.766 & 212.11 $\pm$ 8.44 \\ \cdashline{1-12}
\multirow{3}{*}{\rotatebox[origin=c]{90}{DE}}
& \multicolumn{1}{|l|}{SegFormer} & \textbf{0.784} & 0.033 & 0.887 & 0.416 & 0.798 & - & - & - & - & 667.51 $\pm$ 2.89 \\ 
& \multicolumn{1}{|l|}{DepthFormer}  & - & - & - & - & - & 7.222 & 0.759 & 0.486 & 0.771 & 626.79 $\pm$ 2.05 \\ 
& \multicolumn{1}{|l|}{SegDepthFormer}  & 0.755 & 0.015 & 0.917 & 0.609 & 0.828 & 7.156 & \textbf{0.763} & 0.493 & \textbf{0.773} & 743.23 $\pm$ 32.95 \\ 
\end{tabular}
\end{adjustbox}
\end{center}
\caption{Quantitative comparison on the Cityscapes dataset \cite{cordts2016CityscapesDataset} between the three baseline models paired with MCD, DSE, and DEs, respectively. Best results are marked in \textbf{bold}.}
\label{table: uq cityscapes}
\end{table*}

\begin{table*}[t!]
\begin{center}
\begin{adjustbox}{width=\linewidth}
\setlength\extrarowheight{1mm}
\begin{tabular}{ll|ccccc|cccc|c}
& \multirow{2}{*}{} & \multicolumn{5}{c|}{Semantic Segmentation} & \multicolumn{4}{c|}{Monocular Depth Estimation} & \multirow{2}{*}{Inference Time [ms]}\\ \cdashline{3-11}
& & mIoU $\uparrow$ & ECE $\downarrow$ & p(acc/cer) $\uparrow$ & p(inacc/unc) $\uparrow$ & PAvPU $\uparrow$ & RMSE $\downarrow$ & p(acc/cer) $\uparrow$ & p(inacc/unc) $\uparrow$ & PAvPU $\uparrow$ & \\ \hline \hline
\multirow{3}{*}{\rotatebox[origin=c]{90}{Baseline}}
& \multicolumn{1}{|l|}{SegFormer} & 0.470 & 0.159 & 0.768 & 0.651 & \textbf{0.734} & - & - & - & - & 18.09 $\pm$ 0.41 \\ 
& \multicolumn{1}{|l|}{DepthFormer} & - & - & - & - & - & 0.554 & 0.786 & 0.449 & 0.610 & 17.51 $\pm$ 0.87 \\ 
& \multicolumn{1}{|l|}{SegDepthFormer} & 0.466 & 0.151 & 0.769 & 0.659 & 0.733 & 0.558 & 0.776 & 0.446 & 0.594 & 22.31 $\pm$ 0.23 \\ \cdashline{1-12}
\multirow{3}{*}{\rotatebox[origin=c]{90}{\shortstack{MCD \\ (20\%)}}}
& \multicolumn{1}{|l|}{SegFormer}  & 0.422 & 0.102 & 0.767 & 0.706 & 0.724 & - & - & - & - & 222.67 $\pm$ 0.61 \\ 
& \multicolumn{1}{|l|}{DepthFormer}  & - & - & - & - & - & 0.605 & 0.741 & 0.478 & 0.568 & 139.58 $\pm$ 052 \\ 
& \multicolumn{1}{|l|}{SegDepthFormer}  & 0.433 & 0.093 & 0.771 & 0.710 & 0.725 & 0.610 & 0.731 & 0.450 & 0.560 & 251.25 $\pm$ 0.81 \\ \cdashline{1-12}
\multirow{3}{*}{\rotatebox[origin=c]{90}{\shortstack{MCD \\ (50\%)}}}
& \multicolumn{1}{|l|}{SegFormer}  & 0.273 & 0.083 & 0.705 & \textbf{0.722} & 0.713 & - & - & - & - & 223.25 $\pm$ 0.82 \\ 
& \multicolumn{1}{|l|}{DepthFormer}  & - & - & - & - & - & 0.978 & 0.516 & \textbf{0.492} & 0.526 & 139.27 $\pm$ 0.69 \\ 
& \multicolumn{1}{|l|}{SegDepthFormer}  & 0.272 & 0.084 & 0.702 & 0.721 & 0.711 & 0.837 & 0.576 & 0.473 & 0.525 & 251.98 $\pm$ 0.60 \\ \cdashline{1-12}
\multirow{3}{*}{\rotatebox[origin=c]{90}{\shortstack{DSE}}} 
& \multicolumn{1}{|l|}{SegFormer} & 0.469 & 0.092 & 0.776 & 0.681 & 0.726 & - & - & - & - & 180.42 $\pm$ 3.93 \\ 
& \multicolumn{1}{|l|}{DepthFormer}  & - & - & - & - & - & 0.547 & 0.782 & 0.423 & 0.596 & 91.66 $\pm$ 0.26 \\ 
& \multicolumn{1}{|l|}{SegDepthFormer}  & 0.461 & \textbf{0.077} & 0.776 & 0.692 & 0.723 & 0.584 & 0.738 & 0.403 & 0.573 & 261.69 $\pm$ 5.10 \\ \cdashline{1-12}
\multirow{3}{*}{\rotatebox[origin=c]{90}{\shortstack{DE}}}
& \multicolumn{1}{|l|}{SegFormer}  & \textbf{0.486} & 0.125 & 0.782 & 0.675 & \textbf{0.734} & - & - & - & - & 715.97 $\pm$ 7.55 \\
& \multicolumn{1}{|l|}{DepthFormer}  & - & - & - & - & - & \textbf{0.524} & \textbf{0.808} & 0.475 & \textbf{0.613} & 624.30 $\pm$ 2.07 \\ 
& \multicolumn{1}{|l|}{SegDepthFormer}  & 0.481 & 0.122 & \textbf{0.783} & 0.682 & 0.733 & 0.552 & 0.785 & 0.453 & 0.590 & 788.76 $\pm$ 2.00 \\ 
\end{tabular}
\end{adjustbox}
\end{center}
\caption{Quantitative comparison on the NYUv2 dataset \cite{silberman2012indoor} between the three baseline models paired with MCD, DSE, and DEs, respectively. Best results are marked in \textbf{bold}.}
\label{table: uq nyuv2}
\end{table*}

In this section, we describe the results of our joint uncertainty evaluation quantitatively. We compare combinations of the baseline models SegFormer, DepthFormer, and SegDepthFormer with the uncertainty quantification methods MCD, DSE, and DEs. Tables \ref{table: uq cityscapes} and \ref{table: uq nyuv2} contain a detailed quantitative comparison for the different combinations. The focus particularly lies on the uncertainty quality.

\textbf{Single-task vs. Multi-task.} Looking at the differences between the single-task models, SegFormer and DepthFormer, and the multi-task model, SegDepthFormer, the single-task models generally deliver slightly better prediction performance. However, SegDepthFormer exhibits greater uncertainty quality for the semantic segmentation task in comparison to SegFormer. This is particularly evident for $p(uncertain|inaccurate)$ on Cityscapes. For the depth estimation task, there is no significant difference in terms of uncertainty quality.

\textbf{Baseline Models.} As expected, the baseline models have the lowest inference times, being 5 to 30 times faster without using any uncertainty quantification method. While their prediction performance turns out to be quite competitive, only beaten by DEs, they show poor calibration and uncertainty quality for semantic segmentation. Surprisingly, the uncertainty quality for the depth estimation task is very decent, often only surpassed by the DE. 

\textbf{Monte Carlo Dropout.} The use of MCD causes a significantly higher inference time compared to the respective baseline model. Additionally, leaving dropout activated during inference to sample from the posterior has a detrimental effect on the prediction performance, particularly with a 50\% dropout ratio. Nevertheless, MCD outputs well-calibrated softmax probabilities and uncertainties, although the results should be interpreted with caution because of the deteriorated prediction quality. 

\textbf{Deep Sub-Ensemble.} Across both datasets, DSEs show comparable prediction performance compared with the baseline models. Notably, DSEs consistently demonstrate a high uncertainty quality across all metrics, particularly in the segmentation task on Cityscapes.

\textbf{Deep Ensemble.} In accordance to previous work \cite{ovadia2019DatasetShift, wursthorn2022, gustafsson2020evaluating}, DEs emerge as state-of-the-art, delivering the best prediction performance and mostly superior uncertainty quality. At the same time, DEs suffer from the highest computational cost. 

\section{Efficient Multi-task Uncertainties}\label{sec: emu evaluation}
\begin{table*}[t!]
\begin{center}
\begin{adjustbox}{width=\linewidth}
\setlength\extrarowheight{1mm}
\begin{tabular}{l|ccccc|cccc|c}
\multirow{2}{*}{} & \multicolumn{5}{c|}{Semantic Segmentation} & \multicolumn{4}{c|}{Monocular Depth Estimation} & \multirow{2}{*}{Inference Time [ms]}\\ \cdashline{2-10}
& mIoU $\uparrow$ & ECE $\downarrow$ & p(acc/cer) $\uparrow$ & p(inacc/unc) $\uparrow$ & PAvPU $\uparrow$ & RMSE $\downarrow$ & p(acc/cer) $\uparrow$ & p(inacc/unc) $\uparrow$ & PAvPU $\uparrow$ & \\ \hline \hline
SegDepthFormer (Baseline) & 0.738 & 0.028 & 0.913 & 0.592 & 0.826 & 7.536 & 0.745 & 0.472 & 0.762 & \textbf{22.04 $\pm$ 0.27} \\
SegDepthFormer (DE) & \textbf{0.755} & 0.015 & 0.917 & 0.609 & \textbf{0.828} & 7.156 & 0.763 & \textbf{0.493} & 0.773 & 743.23 $\pm$ 32.95 \\ \cdashline{1-11}
EMUFormer & 0.752 & \textbf{0.012} & \textbf{0.923} & \textbf{0.658} & 0.811 & \textbf{6.983} & \textbf{0.772} & 0.491 & \textbf{0.783} & \textbf{22.04 $\pm$ 0.27} \\
\end{tabular}
\end{adjustbox}
\end{center}
\caption{Quantitative comparison on the Cityscapes dataset \cite{cordts2016CityscapesDataset} between the baseline SegDepthFormer, a SegDepthFormer Deep Ensemble, which acts as the teacher with ten members, and our EMUFormer. Best results are marked in \textbf{bold}.}
\label{table: emu cityscapes}
\end{table*}

\begin{table*}[t!]
\begin{center}
\begin{adjustbox}{width=\linewidth}
\setlength\extrarowheight{1mm}
\begin{tabular}{l|ccccc|cccc|c}
\multirow{2}{*}{} & \multicolumn{5}{c|}{Semantic Segmentation} & \multicolumn{4}{c|}{Monocular Depth Estimation} & \multirow{2}{*}{Inference Time [ms]}\\ \cdashline{2-10}
& mIoU $\uparrow$ & ECE $\downarrow$ & p(acc/cer) $\uparrow$ & p(inacc/unc) $\uparrow$ & PAvPU $\uparrow$ & RMSE $\downarrow$ & p(acc/cer) $\uparrow$ & p(inacc/unc) $\uparrow$ & PAvPU $\uparrow$ & \\ \hline \hline
SegDepthFormer (Baseline) & 0.466 & 0.151 & 0.769 & 0.659 & 0.733 & 0.558 & 0.776 & 0.446 & 0.594 & \textbf{22.31 $\pm$ 0.23} \\ 
SegDepthFormer (DE) & \textbf{0.481} & \textbf{0.122} & 0.783 & 0.682 & 0.733 & 0.552 & 0.785 & \textbf{0.453} & 0.590 & 788.76 $\pm$ 2.00 \\ \cdashline{1-11}
EMUFormer & 0.475 & 0.129 & \textbf{0.787} & \textbf{0.692} & \textbf{0.737} & \textbf{0.514} & \textbf{0.810} & 0.440 & \textbf{0.633} & \textbf{22.31 $\pm$ 0.23}
\end{tabular}
\end{adjustbox}
\end{center}
\caption{Quantitative comparison on the NYUv2 dataset \cite{silberman2012indoor} between the baseline SegDepthFormer, a SegDepthFormer Deep Ensemble, which acts as the teacher with ten members, and our EMUFormer. Best results are marked in \textbf{bold}.}
\label{table: emu nyuv2}
\end{table*}

In this section, we conduct several experiments to demonstrate the efficiency and efficacy of EMUFormer. We begin by comparing EMUFormer's performance with its DE teacher for multiple backbones. Subsequently, we compare our results with previous state-of-the-art approaches, followed by qualitative examples. Lastly, we provide an ablation study on the impact of the GNLL loss. 

\subsection{Quantitative Evaluation}\label{sec: quantitative evaluation}
\begin{table*}[t!]
\begin{center}
\begin{adjustbox}{width=\linewidth}
\setlength\extrarowheight{1mm}
\begin{tabular}{ll|ccccc|cccc|c}
& \multirow{2}{*}{} & \multicolumn{5}{c|}{Semantic Segmentation} & \multicolumn{4}{c|}{Monocular Depth Estimation} & \multirow{2}{*}{Inference Time [ms]}\\ \cdashline{3-11}
& & mIoU $\uparrow$ & ECE $\downarrow$ & p(acc/cer) $\uparrow$ & p(inacc/unc) $\uparrow$ & PAvPU $\uparrow$ & RMSE $\downarrow$ & p(acc/cer) $\uparrow$ & p(inacc/unc) $\uparrow$ & PAvPU $\uparrow$ & \\ \hline \hline
\multirow{4}{*}{\rotatebox[origin=c]{90}{B0 Backbone}}
& \multicolumn{1}{|l|}{SegFormer (DE)} & \textbf{0.689} & 0.037 & 0.888 & 0.486 & 0.779 & - & - & - & - & 273.20 $\pm$ 1.38 \\
& \multicolumn{1}{|l|}{DepthFormer (DE)} & - & - & - & - & - & 8.452 & 0.692 & 0.414 & 0.719 & 236.13 $\pm$ 0.70 \\
& \multicolumn{1}{|l|}{SegDepthFormer (DE)} & 0.651 & 0.045 & 0.912 & 0.634 & \textbf{0.803} & 8.495 & 0.692 & 0.425 & 0.718 & 317.47 $\pm$ 15.64 \\ \cdashline{2-12}
& \multicolumn{1}{|l|}{EMUFormer} & 0.630 & \textbf{0.023} & \textbf{0.924} & \textbf{0.714} & 0.791 & \textbf{8.086} & \textbf{0.717} & \textbf{0.473} & \textbf{0.732} & \textbf{9.58 $\pm$ 0.07} \\ \hline 
\multirow{4}{*}{\rotatebox[origin=c]{90}{\shortstack{B5 Backbone}}}
& \multicolumn{1}{|l|}{SegFormer (DE)} & \textbf{0.809} & 0.032 & 0.896 & 0.435 & 0.819  & - & - & - & - & 1931.01 $\pm$ 12.77 \\ 
& \multicolumn{1}{|l|}{DepthFormer (DE)} & - & - & - & - & - & 6588 & 0.782 & 0.487 & 0.791 & 1892.47 $\pm$ 9.24 \\ 
& \multicolumn{1}{|l|}{SegDepthFormer (DE)} & 0.789 & 0.037 & 0.928 & 0.657 & \textbf{0.852} & 6.664 & 0.785 & 0.502 & 0.792 & 2018.04 $\pm$ 32.31 \\ \cdashline{2-12}
& \multicolumn{1}{|l|}{EMUFormer} & 0.771 & \textbf{0.014} & \textbf{0.934} & \textbf{0.703} & 0.845 & \textbf{6.157} & \textbf{0.804} & \textbf{0.536} & \textbf{0.799} & \textbf{50.72 $\pm$ 0.45} \\
\end{tabular}
\end{adjustbox}
\end{center}
\caption{Quantitative comparison on the Cityscapes dataset \cite{cordts2016CityscapesDataset} between the three baseline models as Deep Ensembles and EMUFormer with SegFormer's B0 and B5 backbone \cite{xie2021segformer}. The respective SegDepthFormer Deep Ensemble served as the teacher for the corresponding EMUFormer. Best results are marked in \textbf{bold}.}
\label{table: architecture cityscapes}
\end{table*}

\begin{table*}[t!]
\begin{center}
\begin{adjustbox}{width=\linewidth}
\setlength\extrarowheight{1mm}
\begin{tabular}{ll|ccccc|cccc|c}
& \multirow{2}{*}{} & \multicolumn{5}{c|}{Semantic Segmentation} & \multicolumn{4}{c|}{Monocular Depth Estimation} & \multirow{2}{*}{Inference Time [ms]}\\ \cdashline{3-11}
& & mIoU $\uparrow$ & ECE $\downarrow$ & p(acc/cer) $\uparrow$ & p(inacc/unc) $\uparrow$ & PAvPU $\uparrow$ & RMSE $\downarrow$ & p(acc/cer) $\uparrow$ & p(inacc/unc) $\uparrow$ & PAvPU $\uparrow$ & \\ \hline \hline
\multirow{4}{*}{\rotatebox[origin=c]{90}{B0 Backbone}}
& \multicolumn{1}{|l|}{SegFormer (DE)} & \textbf{0.376} & 0.105 & 0.743 & 0.701 & 0.718 & - & - & - & - & 315.42 $\pm$ 2.41 \\ 
& \multicolumn{1}{|l|}{DepthFormer (DE)} & - & - & - & - & - & \textbf{0.642} & \textbf{0.720} & 0.476 & \textbf{0.566} & 227.92 $\pm$ 2.39 \\ 
& \multicolumn{1}{|l|}{SegDepthFormer (DE)} & 0.375 & 0.097 & \textbf{0.744} & 0.703 & 0.718 & 0.678 & 0.693 & 0.466 & 0.553 & 346.21 $\pm$ 2.72 \\ \cdashline{2-12}
& \multicolumn{1}{|l|}{EMUFormer} & 0.363 & \textbf{0.090} & 0.743 & \textbf{0.713} & \textbf{0.720} & 0.674 & 0.705 & \textbf{0.498} & 0.558 & \textbf{10.04 $\pm$ 0.06} \\ \hline 
\multirow{4}{*}{\rotatebox[origin=c]{90}{\shortstack{B5 Backbone}}}
& \multicolumn{1}{|l|}{SegFormer (DE)} & \textbf{0.534} & 0.138 & 0.792 & 0.653 & \textbf{0.744} & - & - & - & - & 1958.46 $\pm$ 36.71 \\ 
& \multicolumn{1}{|l|}{DepthFormer (DE)} & - & - & - & - & - & 0.468 & \textbf{0.852} & \textbf{0.505} & \textbf{0.647} & 1875.53 $\pm$ 12.83 \\ 
& \multicolumn{1}{|l|}{SegDepthFormer (DE)} & 0.526 & \textbf{0.133} & 0.794 & 0.665 & 0.743 & \textbf{0.451} & 0.838 & 0.478 & 0.619 & 2038.26 $\pm $13.06 \\ \cdashline{2-12}
& \multicolumn{1}{|l|}{EMUFormer} & 0.520 & 0.134 & \textbf{0.798} & \textbf{0.688} & \textbf{0.744} & 0.476 & 0.846 & 0.467 & \textbf{0.647} & \textbf{52.27 $\pm$ 1.40} \\
\end{tabular}
\end{adjustbox}
\end{center}
\caption{Quantitative comparison on the NYUv2 dataset \cite{silberman2012indoor} between the three baseline models as Deep Ensembles and EMUFormer with SegFormer's B0 and B5 backbone \cite{xie2021segformer}. The respective SegDepthFormer Deep Ensemble served as the teacher for the corresponding EMUFormer. Best results are marked in \textbf{bold}.}
\label{table: architecture nyuv2}
\end{table*}

\begin{table}[th!]
\begin{center}
\begin{adjustbox}{width=\linewidth}
\setlength\extrarowheight{1mm}
\begin{tabular}{l|cc|cc}
& \multicolumn{2}{c|}{NYUv2} & \multicolumn{2}{c}{Cityscapes} \\ \cdashline{2-5}
& mIoU $\uparrow$ & RMSE $\downarrow$ & mIoU $\uparrow$ & RMSE $\downarrow$ \\ \hline \hline
HybridNet A2 \cite{lin2019depth} & 0.343 & 0.682 & 0.666 & 12.09 \\
Mousavian et al. \cite{mousavian2016joint} & 0.392 & 0.816 & - & - \\
C-DCNN \cite{liu2018collaborative} & 0.398 & 0.628 & - & - \\
BMTAS \cite{bruggemann2020automated} & 0.411 & 0.543 & - & - \\
Gao et al. \cite{gao2022predictive} & 0.419 & 0.528 & - & - \\
Nekrasov et al. \cite{nekrasov2019real} & 0.420 & 0.565 & - & - \\
CI-Net \cite{gao2022ci} & 0.426 & 0.504 & 0.701 & 6.880 \\ 
Wang et al. \cite{wang2015towards} & 0.442 & 0.745 & - & - \\
SOSD-Net \cite{he2021sosd} & 0.450 & 0.514 & 0.682 & - \\ 
ATRC \cite{bruggemann2021exploring} & 0.463 & 0.536 & - & - \\ 
MTI-Net \cite{vandenhende2020mti} & 0.490 & 0.529 & - & - \\ 
PAD-Net \cite{xu2018pad} & 0.502 & 0.582 & 0.761 & - \\ 
MTFormer \cite{xu2022mtformer} & 0.506 & \underline{0.483} & - & - \\ \hline 
SegDepthFormer-B2 (Ours) & 0.476 & 0.549 & 0.763 & 7.286 \\
SegDepthFormer-B5 (Ours) & \underline{0.518} & 0.499 & \textbf{0.784} & \underline{6.819} \\ \cdashline{1-5}
EMUFormer-B2 (Ours) & 0.475 & 0.514 & 0.752 & 6.983 \\
EMUFormer-B5 (Ours) & \textbf{0.520} & \textbf{0.476} & \underline{0.771} & \textbf{6.157} \\
\end{tabular}
\end{adjustbox}
\end{center}
\caption{Comparison against previous state-of-the-art approaches in joint semantic segmentation and monocular depth estimation. Best results are marked in \textbf{bold}, second best results are \underline{underlined}.}
\label{table: sota}
\end{table}

\textbf{Baseline vs. Teacher vs. Student.} We present a comprehensive analysis in Tables \ref{table: emu cityscapes} and \ref{table: emu nyuv2} by comparing SegDepthFormer (baseline), SegDepthFormer DE (teacher), and EMUFormer (student). EMUFormer emerges as the standout performer, surpassing the baseline SegDepthFormer model across all metrics on both datasets, with only a single exception. Remarkably, this performance is achieved while maintaining an equivalent inference time. Remarkably, EMUFormer even outperforms the SegDepthFormer DE, which served as its teacher and has approximately 33 times higher inference time, in most cases. In terms of prediction performance, EMUFormer gives slightly worse segmentation results compared to the DE. However, it notably excels in the depth estimation task, especially on Cityscapes \cite{cordts2016CityscapesDataset}, which is a phenomenon we observed across multiple experiments (cf. Tables \ref{table: architecture cityscapes}, \ref{table: architecture nyuv2}, and \ref{table: sota}) and which we will discuss in Section \ref{sec: discussion}.

\textbf{Backbone Size.} Tables \ref{table: architecture cityscapes} and \ref{table: architecture nyuv2} display a comprehensive assessment of the influence of the backbone size on Cityscapes \cite{cordts2016CityscapesDataset} and NYUv2 \cite{silberman2012indoor}. In this context, we decided to evaluate the three baseline models as a DE with ten members each in comparison to EMUFormer for the smallest, B0, and the biggest, B5, backbone of SegFormer \cite{xie2021segformer}, respectively. The findings broadly align with the earlier observations of Section \ref{sec: joint uncertainty evaluation} in terms of single-tasking versus multi-tasking. More specifically, EMUFormer emerges as the top performer on all segmentation metrics, except for the mIoU where the SegFormer DE gives slightly better results. On the Cityscapes dataset, EMUFormer stands out by delivering the best results for all depth metrics across both backbones. Notably, it achieves this superior performance while maintaining a 20 to 30 times faster inference time compared to the DEs. On NYUv2, the DepthFormer DE performs marginally better on the depth metrics, although EMUFormer remains highly competitive, especially if inference time is considered.  

\textbf{Comparison with SOTA.} 
On both datasets, Cityscapes \cite{cordts2016CityscapesDataset} and NYUv2 \cite{silberman2012indoor}, EMUFormer-B5 outperforms the previous state-of-the-art in joint semantic segmentation and monocular depth estimation. For instance, on NYUv2 \cite{silberman2012indoor}, EMUFormer delivers 1.4\,\% higher mIoU and 0.007 lower RMSE than MTFormer \cite{xu2022mtformer}, which also adopts a modern Vision-Transformer-based architecture. In contrast to our work, however, they rely on cross-task attention mechanisms and on a sophisticated self-supervised pre-training routine, which introduce additional complexity. Our SegDepthFormer-B5 baseline model already achieves very competitive results without such adaptations to the architecture or the training routine. It improves upon previous work in all cases except for RMSE on NYUv2 \cite{silberman2012indoor}. Besides, even with the lightweight B2 models, we achieve very decent results in comparison to prior work, offering an alternative for real-time applications.

\subsection{Qualitative Evaluation}
\begin{figure*}
    \centering

    \begin{tabular}{cccc}
        \begin{subfigure}{0.24\textwidth}
            \includegraphics[width=\textwidth]{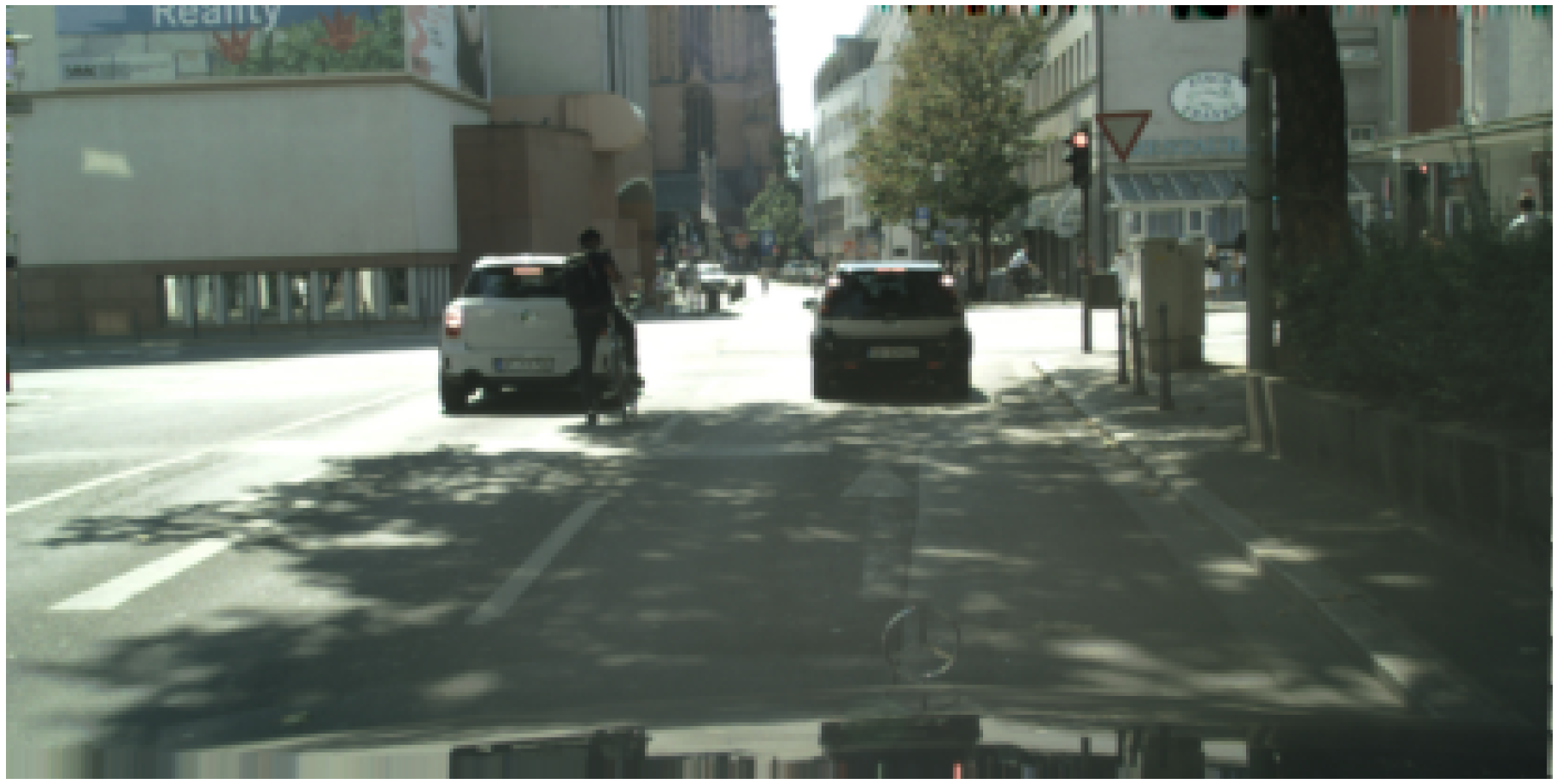}
            \caption{Input Image}
        \end{subfigure} &
        \begin{subfigure}{0.24\textwidth}
            \includegraphics[width=\textwidth]{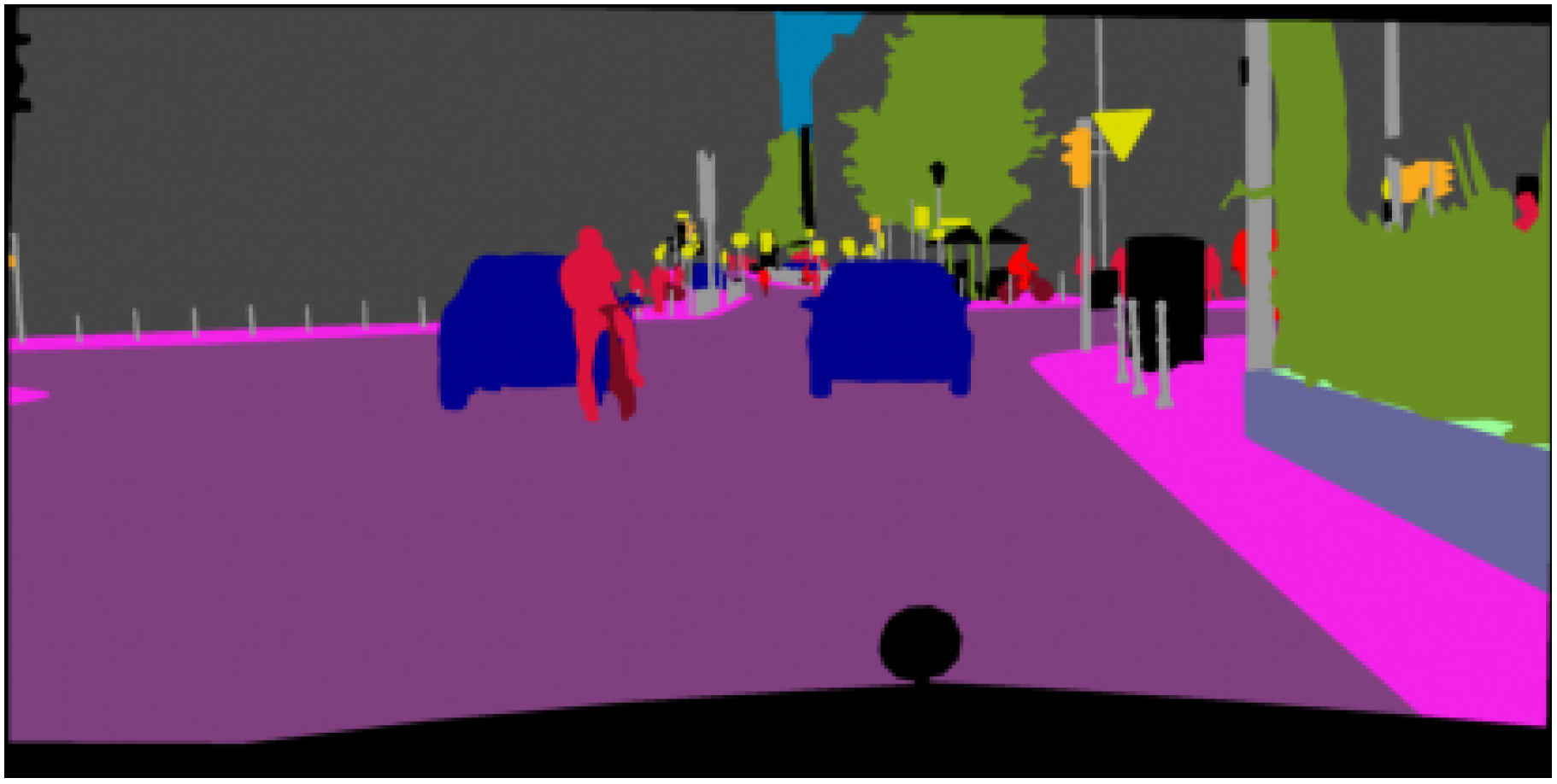}
            \caption{Segmentation Ground Truth}
            \label{subfig:city_seg_gt}
        \end{subfigure} &
        \begin{subfigure}{0.24\textwidth}
            \includegraphics[width=\textwidth]{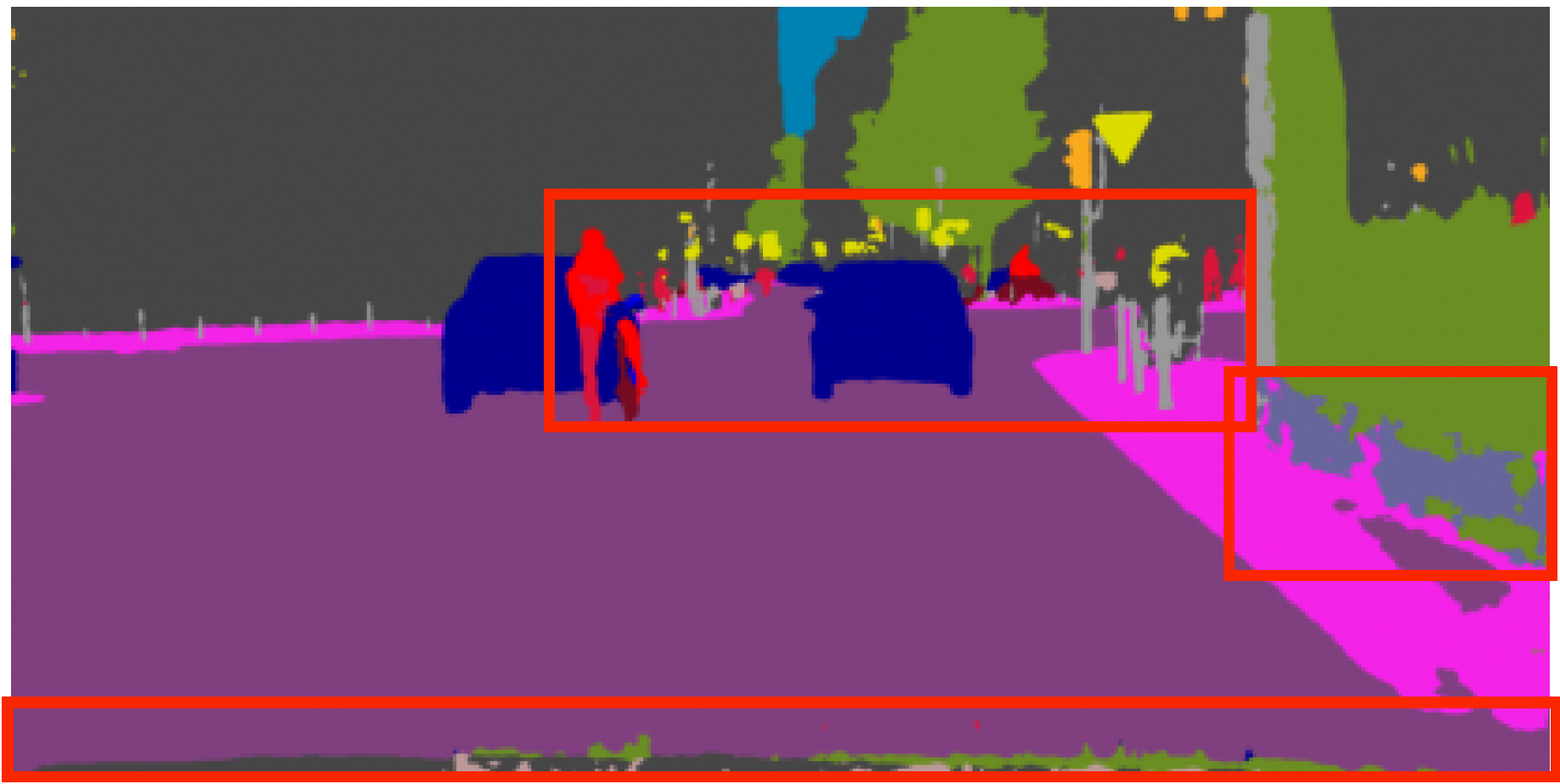}
            \caption{Segmentation Prediction}
            \label{subfig:city_seg_pred}
        \end{subfigure} &
        \begin{subfigure}{0.24\textwidth}
            \includegraphics[width=\textwidth]{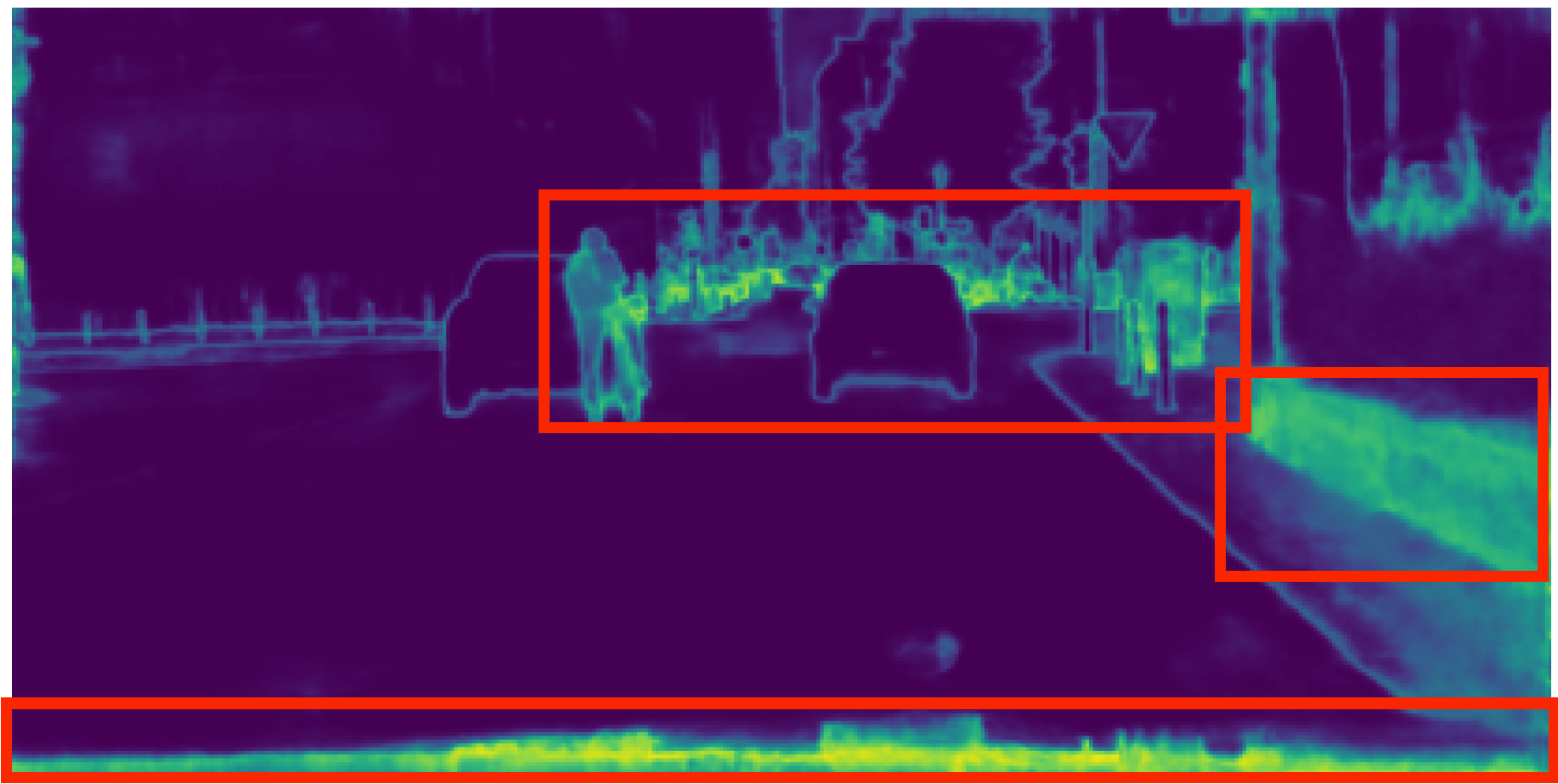}
            \caption{Segmentation Uncertainty}
            \label{subfig:city_seg_unc}
        \end{subfigure} \\
        & 
        \begin{subfigure}{0.24\textwidth}
            \includegraphics[width=\textwidth]{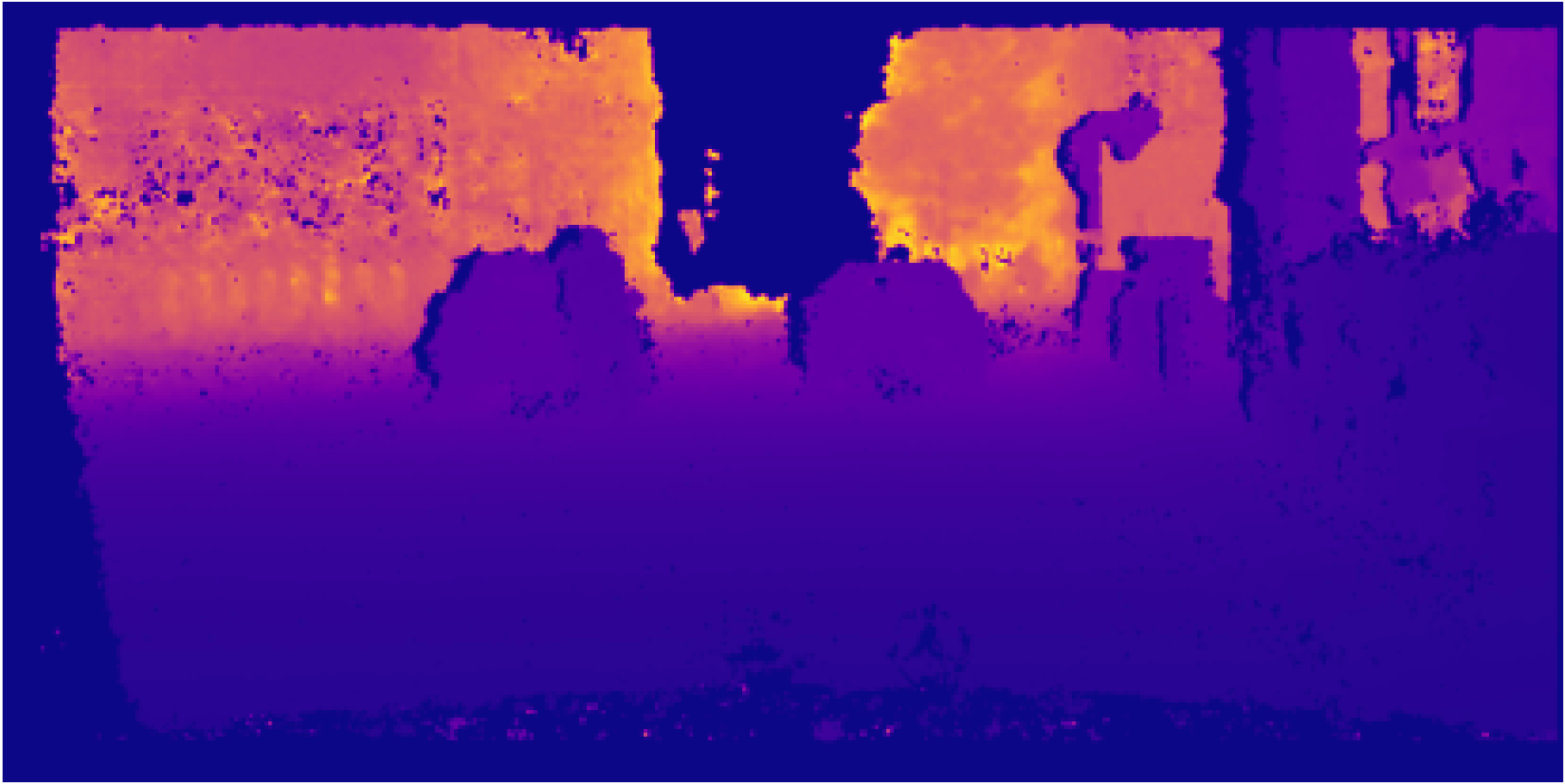}
            \caption{Depth Ground Truth}
        \end{subfigure} &
        \begin{subfigure}{0.24\textwidth}
            \includegraphics[width=\textwidth]{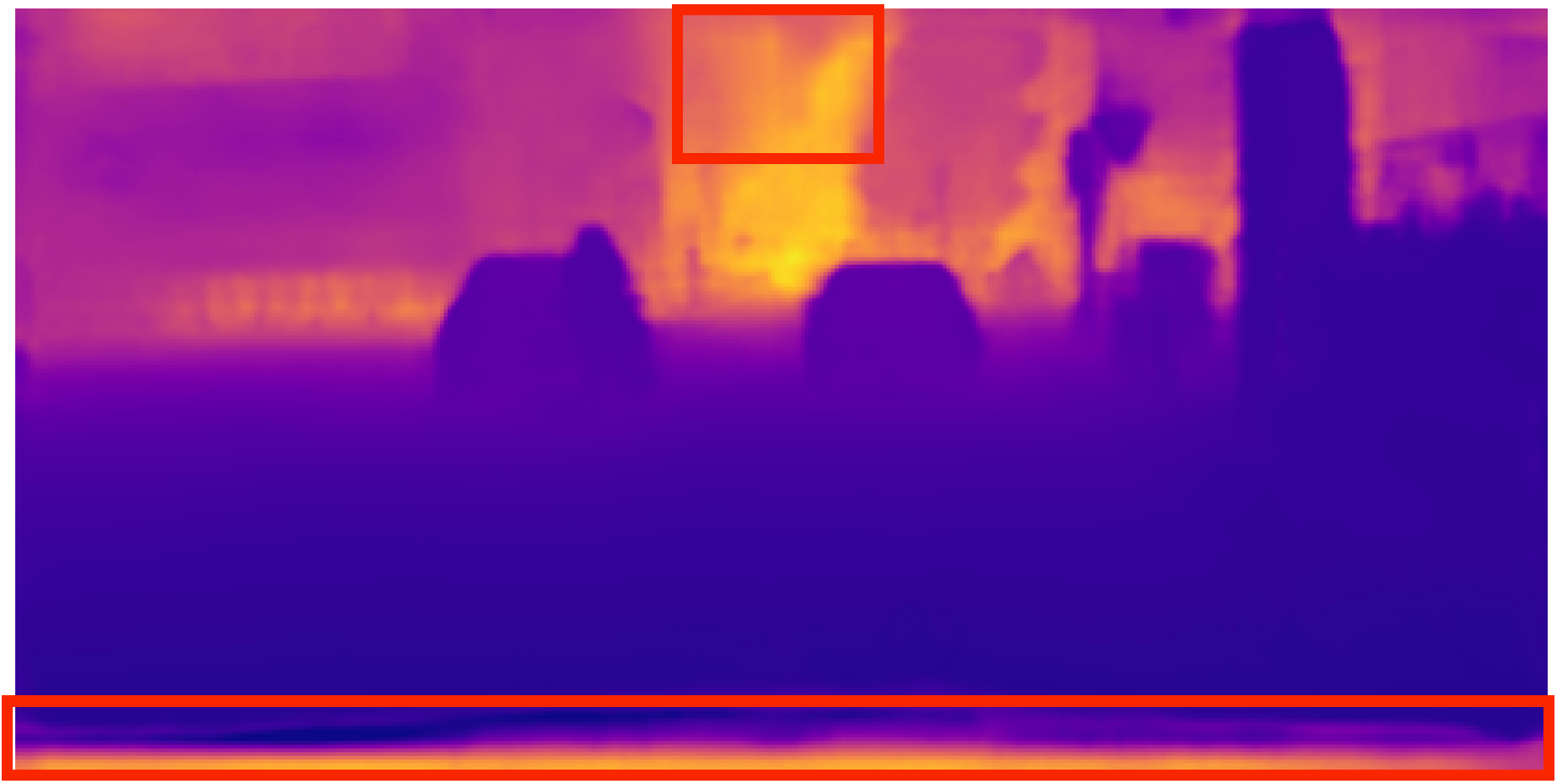}
            \caption{Depth Prediction}
        \end{subfigure} &
        \begin{subfigure}{0.24\textwidth}
            \includegraphics[width=\textwidth]{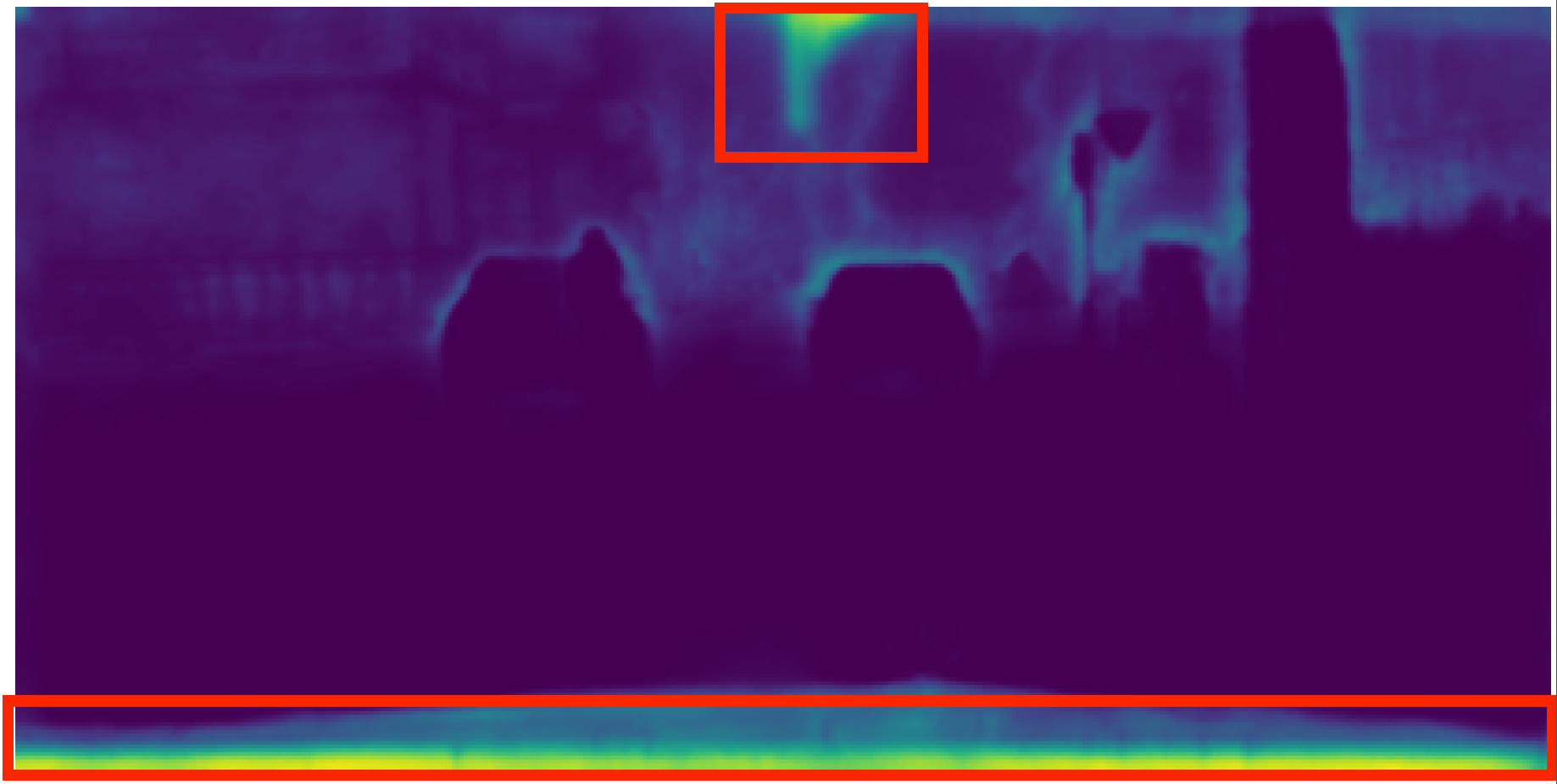}
            \caption{Depth Uncertainty}
        \end{subfigure} \\
        
        \begin{subfigure}{0.24\textwidth}
            \includegraphics[width=\textwidth]{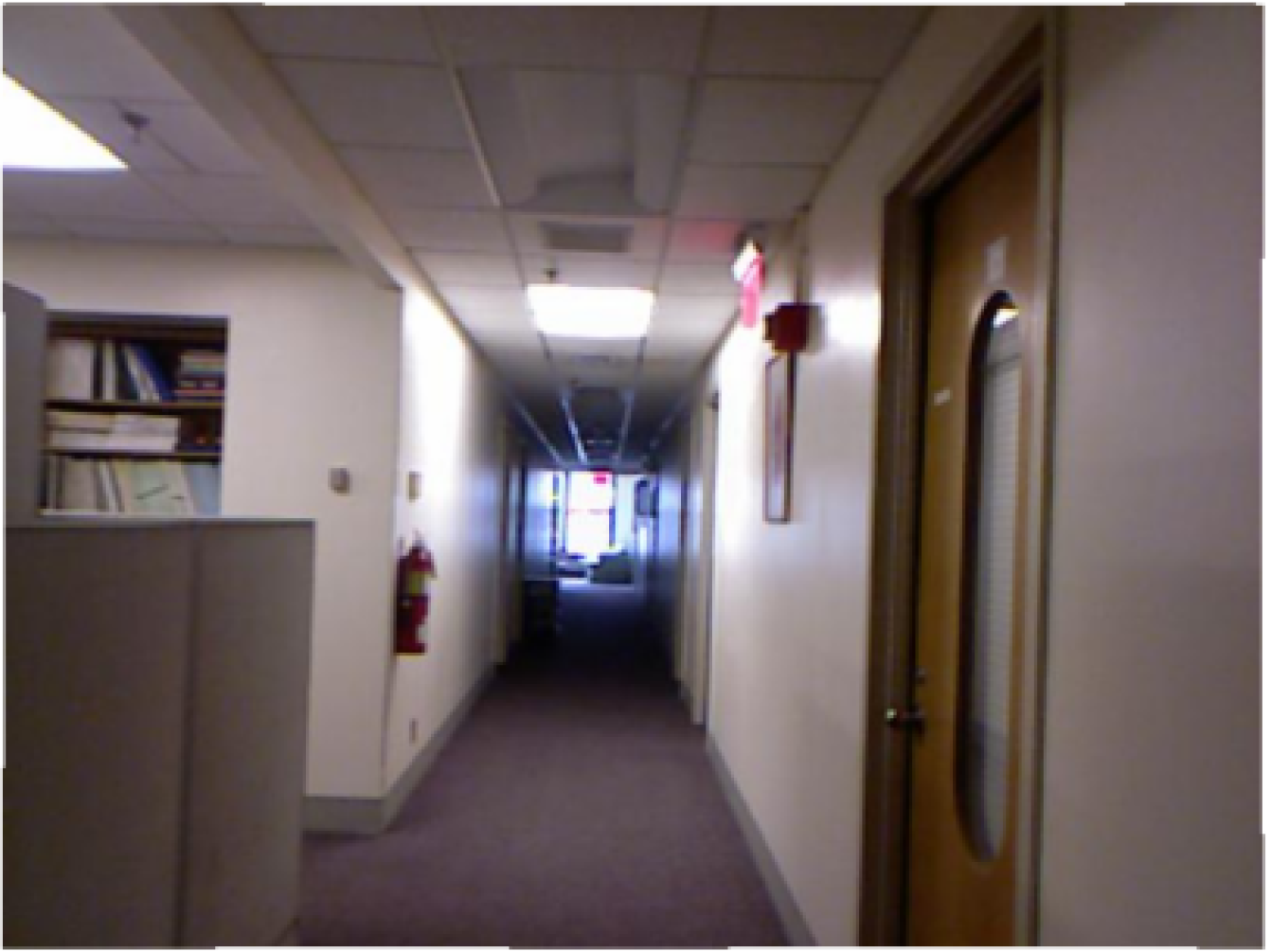}
            \caption{Input Image}
        \end{subfigure} &
        \begin{subfigure}{0.24\textwidth}
            \includegraphics[width=\textwidth]{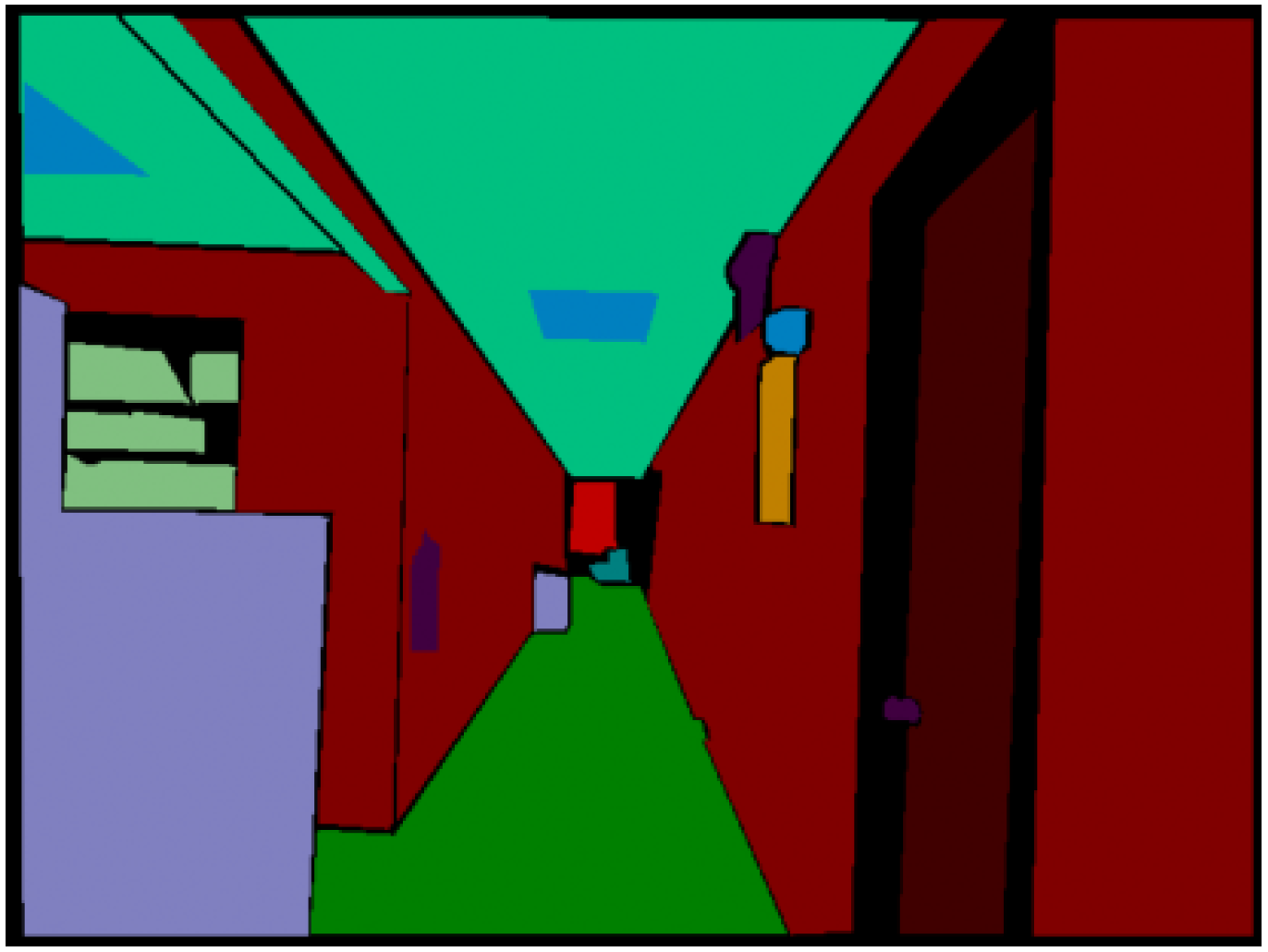}
            \caption{Segmentation Ground Truth}
            \label{subfig:nyuv2_seg_gt}
        \end{subfigure} &
        \begin{subfigure}{0.24\textwidth}
            \includegraphics[width=\textwidth]{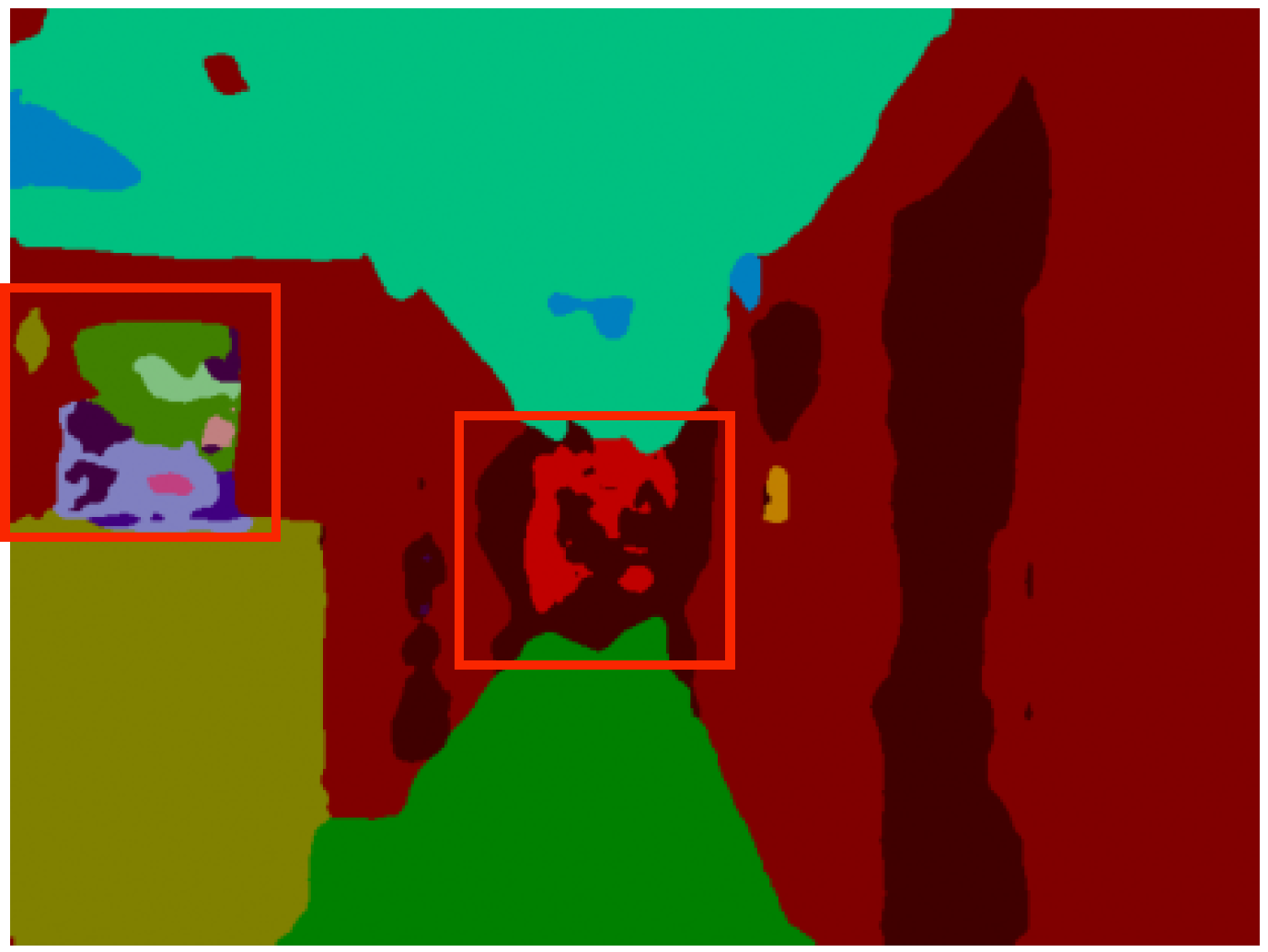}
            \caption{Segmentation Prediction}
            \label{subfig:nyuv2_seg_pred}
        \end{subfigure} &
        \begin{subfigure}{0.24\textwidth}
            \includegraphics[width=\textwidth]{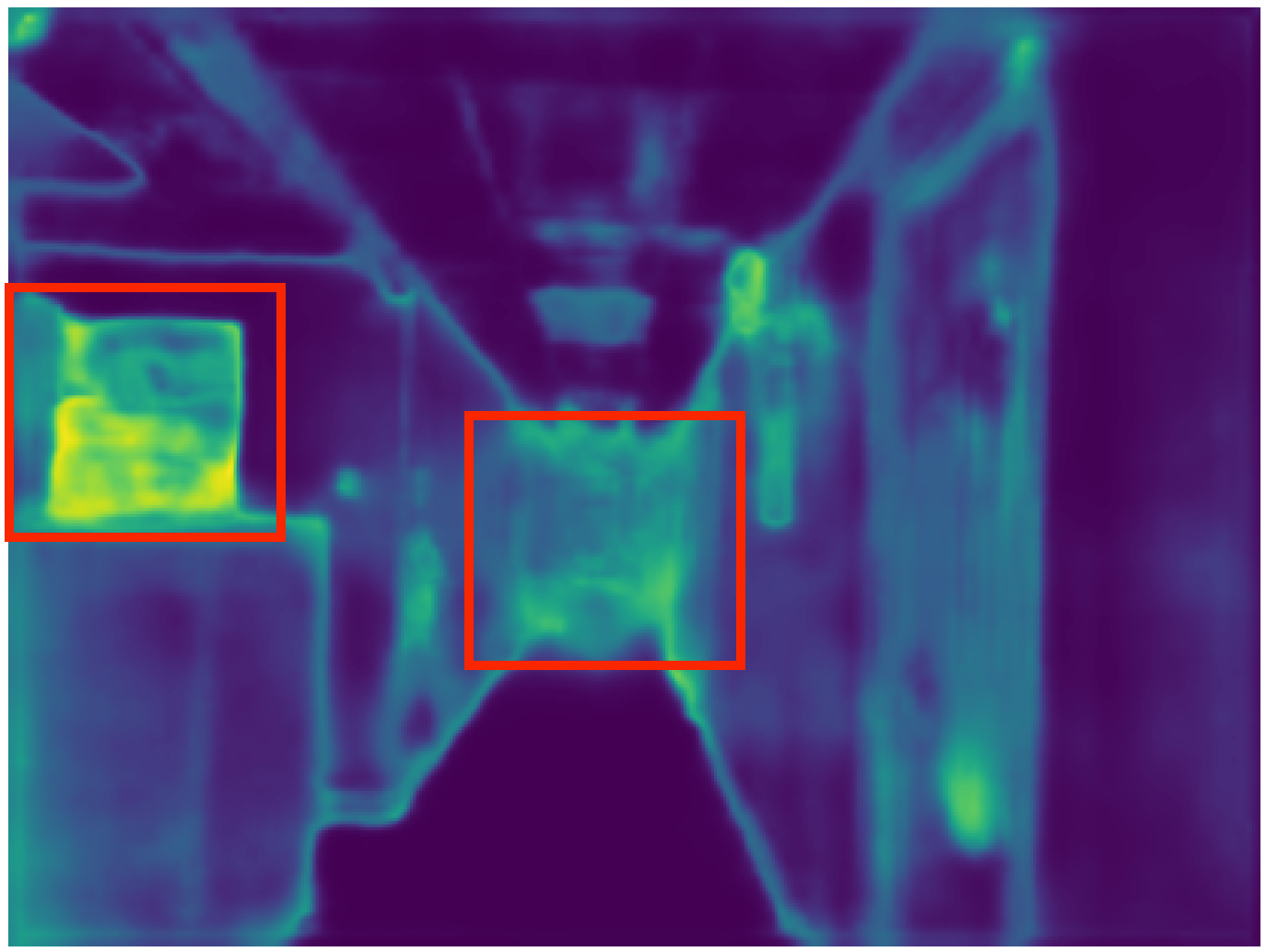}
            \caption{Segmentation Uncertainty}
            \label{subfig:nyuv2_seg_unc}
        \end{subfigure} \\
        & 
        \begin{subfigure}{0.24\textwidth}
            \includegraphics[width=\textwidth]{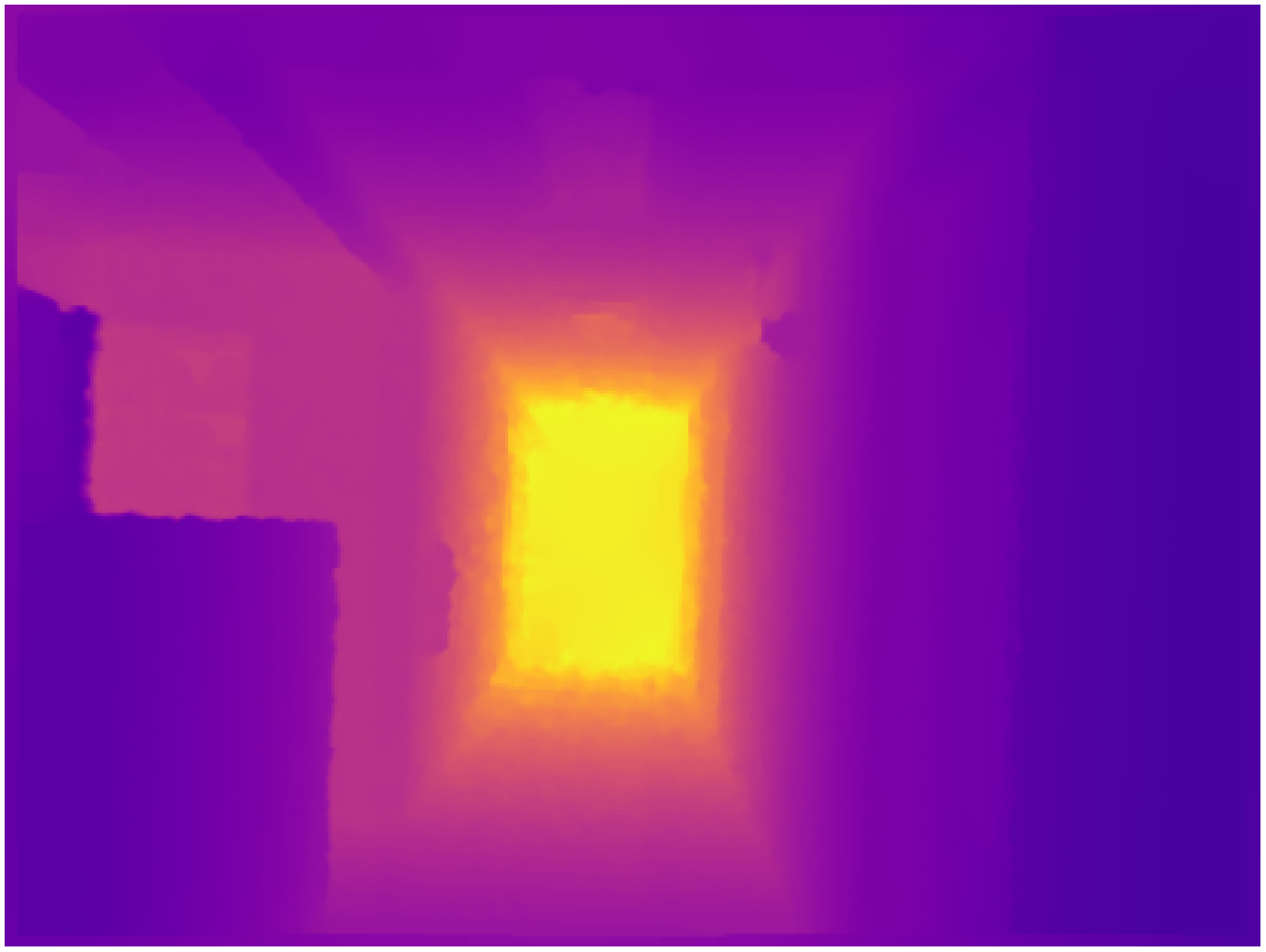}
            \caption{Depth Ground Truth}
        \end{subfigure} &
        \begin{subfigure}{0.24\textwidth}
            \includegraphics[width=\textwidth]{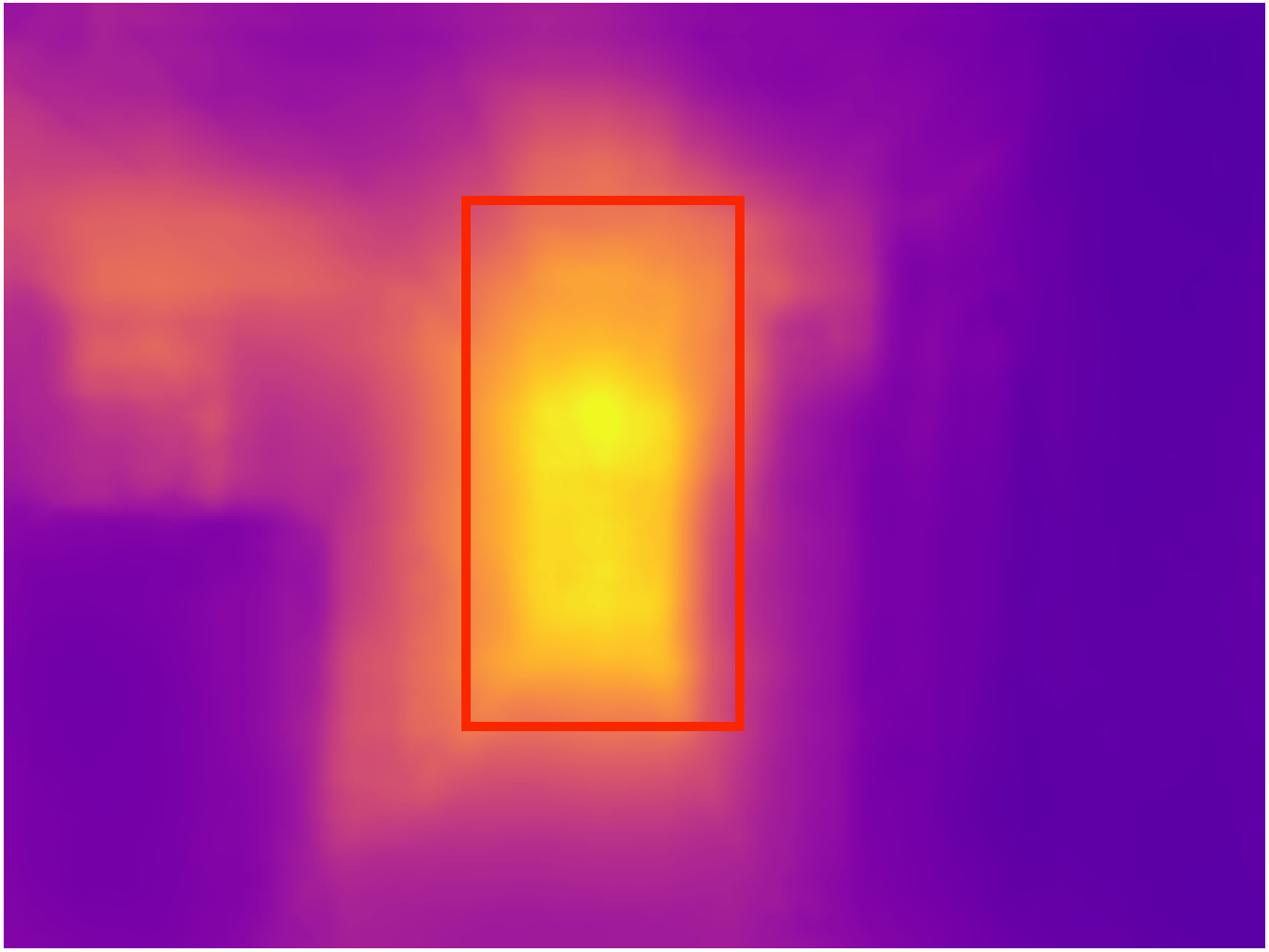}
            \caption{Depth Prediction}
        \end{subfigure} &
        \begin{subfigure}{0.24\textwidth}
            \includegraphics[width=\textwidth]{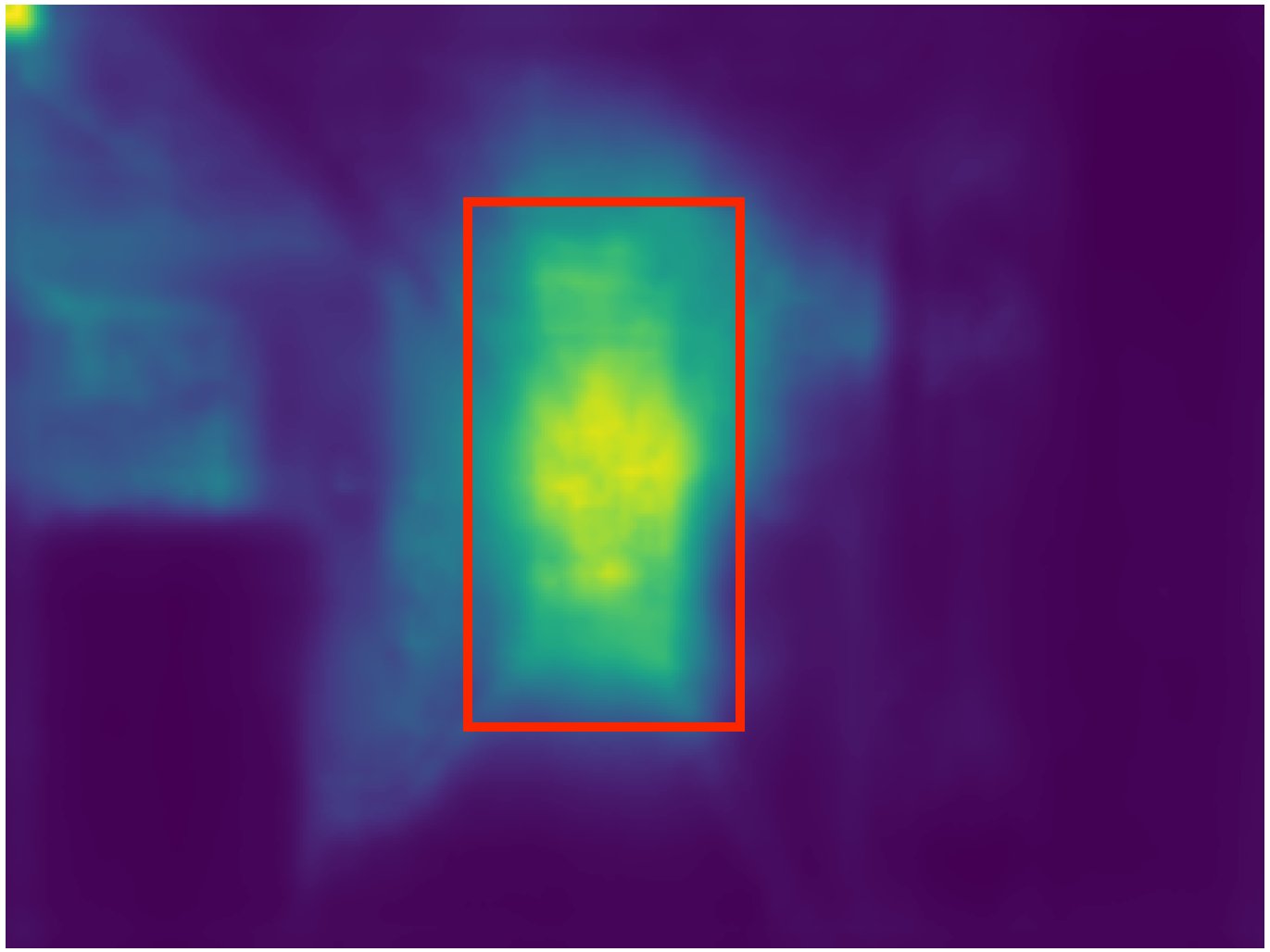}
            \caption{Depth Uncertainty}
        \end{subfigure} \\
    \end{tabular}

    \caption{Qualitative examples of our EMUFormer-B2 on the Cityscapes \cite{cordts2016CityscapesDataset} (top) and NYUv2 \cite{silberman2012indoor} (bottom) datasets. Red rectangles are added to highlight interesting areas. Best viewed in color.}

    \label{fig:qualitative_examples}

\end{figure*}

In addition to the quantitative evaluation, we also provide qualitative examples of EMUFormer-B2 in Figure \ref{fig:qualitative_examples} for Cityscapes \cite{cordts2016CityscapesDataset} and NYUv2 \cite{silberman2012indoor}.

\textbf{Cityscapes.} On Cityscapes, EMUFormer demonstrates good prediction performance for both tasks. In the segmentation task, its uncertainty prediction proves particularly insightful as the red rectangles highlight. For example, in areas such as the car hood, which is not part of the training labels (indicated by black pixels), the model exhibits high uncertainty, indicating its ability to capture out-of-distribution information or epistemic uncertainty. Similarly, in noisy background areas, the model effectively captures the aleatoric noise. Additionally, the model correctly predicts high uncertainties for challenging areas like the wall on the right of the image, highlighting the utility of uncertainties in identifying potential model errors. In the depth estimation task, analogous to the segmentation task, EMUFormer predicts high uncertainty on the car hood or the sky, which are both areas that are not part of the training ground truth, i.e. areas of high epistemic uncertainty. Furthermore, the uncertainty is appropriately high at object boundaries, indicating sensitivity to significant depth variations.

\textbf{NYUv2.} 
For the segmentation task, EMUFormer again outputs high uncertainties for pixels that are not part of the ground truth or those that are misclassified, consistently providing useful predictive uncertainties. In the depth estimation task, the uncertainties seem to correlate with the estimated depth, providing an intuitive and helpful indication. This alignment suggests that the model effectively captures the depth prediction quality, particularly as it relates to increasing distances.

In summary, the qualitative evaluation aligns with the quantitative findings of Section \ref{sec: quantitative evaluation} and highlights EMUFormer's proficiency in handling both the segmentation and the depth estimation tasks, showcasing its ability to generate meaningful predictive uncertainties that enable more thorough interpretations of the predictions.

\subsection{Ablation Studies}\label{sec:abl}
\textbf{Impact of GNLL Loss.} EMUFormer is trained to minimize the weighted sum of the following four objective functions:
\begin{enumerate}
    \itemsep0em
    \item Cross-Entropy loss for the semantic segmentation task.
    \item Kullback-Leibler divergence loss for the segmentation uncertainty distillation.
    \item Gaussian Negative Log-Likelihood loss for the monocular depth estimation task.
    \item Root mean squared error loss for the depth uncertainty distillation.
\end{enumerate}
As described in Section \ref{sec: depthformer} and shown by Equation \ref{eq: gnll}, GNLL treats every prediction as a sample from a Gaussian distribution with a predictive mean and a corresponding predictive variance. Usually, these variances are solely learned implicitly through the optimization of the predictive means based on the ground truth labels. In the case of EMUFormer, however, the network is also being trained to mimic the predictive uncertainty of the teacher in parallel. Consequently, the depth uncertainty does not need to be learned implicitly, rather it can be used to improve the depth estimation itself. In order to explore this more thoroughly, we performed an ablation study on the impact of the GNLL loss by replacing the GNLL loss with the Mean Squared Error (MSE) loss and the Huber loss \cite{huber1992robust}, respectively.

Tables \ref{table: gnll ablation study cityscapes} and \ref{table: gnll ablation study nyuv2} show a quantitative comparison of the impact of the respective depth loss for EMUFormer-B2 on the Cityscapes and NYUv2 datasets. On Cityscapes, training with GNLL loss leads to the best performance across the board, especially with regard to the RMSE for monocular depth estimation. GNLL loss results in a RMSE of 6.983 in comparison to 7.217 and 7.340 for MSE and Huber loss \cite{huber1992robust}, respectively. Similarly, on NYUv2, training with GNLL loss yields the best RMSE with 0.514 versus 0.527 and 0.533 for MSE and Huber loss \cite{huber1992robust}, although at the cost of a very slight deterioration of 0.006 in mIoU. Remarkably, using GNLL loss leads to the highest depth uncertainty quality for both datasets. 

Overall, these results show that incorporating the predictive uncertainties using the GNLL loss enhances the performance of EMUFormer for depth estimation and depth uncertainty quantification compared to other loss functions like MSE or Huber loss \cite{huber1992robust} that do not account for the uncertainty. We consider this a valuable insight and believe that leveraging high-quality predictive uncertainties during the optimization process offers great potential for future work.

\begin{table*}[t!]
\begin{center}
\begin{adjustbox}{width=\linewidth}
\setlength\extrarowheight{1mm}
\begin{tabular}{l|ccccc|cccc|c}
\multirow{2}{*}{} & \multicolumn{5}{c|}{Semantic Segmentation} & \multicolumn{4}{c|}{Monocular Depth Estimation} \\ \cdashline{2-10}
& mIoU $\uparrow$ & ECE $\downarrow$ & p(acc/cer) $\uparrow$ & p(inacc/unc) $\uparrow$ & PAvPU $\uparrow$ & RMSE $\downarrow$ & p(acc/cer) $\uparrow$ & p(inacc/unc) $\uparrow$ & PAvPU $\uparrow$ \\ \hline \hline
MSE & 0.749 & 0.014 & 0.922 & \textbf{0.659} & 0.810 & 7.217 & 0.742 & 0.446 & 0.761 \\ 
Huber \cite{huber1992robust} & 0.748 & 0.013 & \textbf{0.923} & 0.657 & 0.809 & 7.340 & 0.743 & 0.446 & 0.760 \\ \cdashline{1-10}
GNLL & \textbf{0.752} & \textbf{0.012} & \textbf{0.923} & 0.658 & \textbf{0.811} & \textbf{6.983} & \textbf{0.772} & \textbf{0.491} & \textbf{0.783}
\end{tabular}
\end{adjustbox}
\end{center}
\caption{Ablation study on the impact of the depth loss on the results of EMUFormer-B2 on Cityscapes \cite{cordts2016CityscapesDataset}. Best results are marked in \textbf{bold}.}
\label{table: gnll ablation study cityscapes}
\end{table*}

\begin{table*}[t!]
\begin{center}
\begin{adjustbox}{width=\linewidth}
\setlength\extrarowheight{1mm}
\begin{tabular}{l|ccccc|cccc|c}
\multirow{2}{*}{} & \multicolumn{5}{c|}{Semantic Segmentation} & \multicolumn{4}{c|}{Monocular Depth Estimation} \\ \cdashline{2-10}
& mIoU $\uparrow$ & ECE $\downarrow$ & p(acc/cer) $\uparrow$ & p(inacc/unc) $\uparrow$ & PAvPU $\uparrow$ & RMSE $\downarrow$ & p(acc/cer) $\uparrow$ & p(inacc/unc) $\uparrow$ & PAvPU $\uparrow$ \\ \hline \hline
MSE & \textbf{0.481} & \textbf{0.127} & \textbf{0.788} & 0.690 & \textbf{0.737} & 0.527 & 0.788 & 0.431 & 0.587 \\ 
Huber \cite{huber1992robust} & \textbf{0.481} & \textbf{0.127} & \textbf{0.788} & 0.689 & \textbf{0.737} & 0.533 & 0.786 & 0.431 & 0.587 \\ \cdashline{1-10}
GNLL & 0.475 & 0.129 & 0.787 & \textbf{0.692} & \textbf{0.737} & \textbf{0.514} & \textbf{0.810} & \textbf{0.440} & \textbf{0.633}
\end{tabular}
\end{adjustbox}
\end{center}
\caption{Ablation study on the impact of the depth loss on the results of EMUFormer-B2 on NYUv2 \cite{silberman2012indoor}. Best results are marked in \textbf{bold}.}
\label{table: gnll ablation study nyuv2}
\end{table*}

\section{Discussion}\label{sec: discussion}
\textbf{Joint Uncertainty Evaluation.} Quantifying the uncertainty in joint segmentation and depth estimation has not been thoroughly examined in prior research. Therefore, we evaluated multiple uncertainty quantification methods with modern Vision-Transformer-based architectures for joint semantic segmentation and monocular depth estimation. In general, single-task models demonstrate slightly better prediction performance, which may arise from multiple factors. For one, the single-task models can optimize all available parameters for their specific task. Additionally, the multi-task models do not exploit any sophisticated adaptions to the architecture or the training process. Unlike previous work \cite{wang2015towards,mousavian2016joint,jiao2018look,xu2018pad,liu2018collaborative,lin2019depth,nekrasov2019real,he2021sosd,gao2022ci,ji2023semantic,kendall2018multi,liu2019end,bruggemann2021exploring,xu2022mtformer,bruggemann2020automated}, we intentionally left out all of the complexities for the joint uncertainty evaluation in order to maintain methodological simplicity and transparency of the results. Interestingly, multi-task models showcase greater uncertainty quality, particularly in the context of the semantic segmentation task. This suggests that jointly training a model to solve multiple tasks can enhance the model's ability to better quantify its uncertainty. In terms of uncertainty quantification methods, DEs stand out as the preferred choice, demonstrating superior prediction performance and, for the most part, higher uncertainty quality. However, it is crucial to note that this advantage comes at the highest computational cost. Both findings align closely with previous work focusing on the evaluation of uncertainties \cite{ovadia2019DatasetShift, wursthorn2022, gustafsson2020evaluating}. Among the more efficient methods, MCD and DSE, the latter exhibits a prediction performance that is comparable with the baseline models while achieving a high uncertainty quality. This positions DSEs as an attractive alternative to DEs, offering efficiency without significant sacrifices in performance or uncertainty quality.  

\textbf{EMUFormer.} In addition to the joint uncertainty evaluation, we also proposed EMUFormer, which employs student-teacher distillation to achieve state-of-the-art results in joint semantic segmentation and monocular depth estimation on Cityscapes \cite{cordts2016CityscapesDataset} and NYUv2 \cite{silberman2012indoor}. Notably, it accomplishes this while estimating well-calibrated predictive uncertainties for both tasks, all without introducing any additional computational overhead during inference. Remarkably, EMUFormer even surpasses the performance of its DE teacher in certain cases, despite the latter having ten times the parameters and approximately 30 times higher inference time. The backbone ablation analysis further reinforces the versatility of our proposed method, showcasing its efficacy across different backbone configurations. Most interestingly, however, EMUFormer achieves particularly outstanding performance in the depth estimation task in comparison to the teacher. We primarily attribute this success to the use of the Gaussian Negative Log-Likelihood loss (cf. Section \ref{sec:abl}), which is commonly employed to implicitly learn corresponding variances in addition to the predictive means. In the case of EMUFormer, however, the teacher model already provides high-quality variances through distillation, allowing for a more accurate approximation of the predictive means and their associated uncertainties. Consequently, leveraging uncertainties during the training, either implicitly like EMUFormer, or explicitly like previous work \cite{landgraf2023u, kendall2018multi}, is an interesting venue for future work.

\section{Conclusion}\label{sec: conclusion}
In this work, we first combine multiple uncertainty quantification methods with joint semantic segmentation and monocular depth estimation and evaluate how they perform in comparison to each other. Quantitative evaluations revealed that Deep Ensembles stand out as the preferred choice concerning prediction performance and uncertainty quality, although having the highest computational cost. Among the less costly methods, Deep Sub-Ensembles emerge as an attractive alternative to Deep Sub-Ensembles, offering efficiency without major sacrifices in prediction performance or uncertainty quality. Additionally, we reveal the benefits of multi-task learning with regard to the uncertainty quality compared to solving both tasks separately. Building on these insights, we propose EMUFormer, a novel student-teacher distillation approach for joint semantic segmentation and monocular depth estimation as well as efficient multi-task uncertainty quantification. By implicitly leveraging the predictive uncertainties of the teacher, EMUFormer achieves new state-of-the-art results on Cityscapes and NYUv2 for both tasks. Notably, EMUFormer also manages to estimate high-quality predictive uncertainties for both tasks that are comparable or superior to a DE despite being an order of magnitude more efficient.

\section*{Acknowledgment}
The authors acknowledge support by the state of Baden-Württemberg through bwHPC.

This work is supported by the Helmholtz Association Initiative and Networking Fund on the HAICORE@KIT partition.

{\small
\bibliographystyle{ieee_fullname}
\bibliography{literature}

\begin{thebibliography}{10}\itemsep=-1pt

\bibitem{amini2020deep}
Alexander Amini, Wilko Schwarting, Ava Soleimany, and Daniela Rus.
\newblock Deep evidential regression.
\newblock {\em Advances in Neural Information Processing Systems}, 33:14927--14937, 2020.

\bibitem{besnier2021learning}
Victor Besnier, David Picard, and Alexandre Briot.
\newblock Learning uncertainty for safety-oriented semantic segmentation in autonomous driving.
\newblock In {\em 2021 IEEE International Conference on Image Processing (ICIP)}, pages 3353--3357. IEEE, 2021.

\bibitem{bruggemann2020automated}
David Bruggemann, Menelaos Kanakis, Stamatios Georgoulis, and Luc Van~Gool.
\newblock Automated search for resource-efficient branched multi-task networks.
\newblock {\em arXiv preprint arXiv:2008.10292}, 2020.

\bibitem{bruggemann2021exploring}
David Br{\"u}ggemann, Menelaos Kanakis, Anton Obukhov, Stamatios Georgoulis, and Luc Van~Gool.
\newblock Exploring relational context for multi-task dense prediction.
\newblock In {\em Proceedings of the IEEE/CVF international conference on computer vision}, pages 15869--15878, 2021.

\bibitem{LiangfuDriving}
Liangfu Chen, Zeng Yang, Jianjun Ma, and Zheng Luo.
\newblock Driving scene perception network: Real-time joint detection, depth estimation and semantic segmentation.
\newblock In {\em 2018 IEEE Winter Conference on Applications of Computer Vision (WACV)}, pages 1283--1291, 2018.

\bibitem{cordts2016CityscapesDataset}
Marius Cordts, Mohamed Omran, Sebastian Ramos, Timo Rehfeld, Markus Enzweiler, Rodrigo Benenson, Uwe Franke, Stefan Roth, and Bernt Schiele.
\newblock The cityscapes dataset for semantic urban scene understanding.
\newblock In {\em Proceedings of the IEEE Conference on Computer Vision and Pattern Recognition (CVPR)}, June 2016.

\bibitem{Deng_2021_ICCV}
Didan Deng, Liang Wu, and Bertram~E. Shi.
\newblock Iterative distillation for better uncertainty estimates in multitask emotion recognition.
\newblock In {\em Proceedings of the IEEE/CVF International Conference on Computer Vision (ICCV) Workshops}, pages 3557--3566, October 2021.

\bibitem{deng2009ImageNetLargescale}
Jia Deng, Wei Dong, Richard Socher, Li-Jia Li, {Kai Li}, and {Li Fei-Fei}.
\newblock {{ImageNet}}: {{A}} large-scale hierarchical image database.
\newblock In {\em 2009 {{IEEE Conference}} on {{Computer Vision}} and {{Pattern Recognition}}}, pages 248--255, {Miami, FL}, 2009. {IEEE}.

\bibitem{dong2022towards}
Xingshuai Dong, Matthew~A Garratt, Sreenatha~G Anavatti, and Hussein~A Abbass.
\newblock Towards real-time monocular depth estimation for robotics: A survey.
\newblock {\em IEEE Transactions on Intelligent Transportation Systems}, 23(10):16940--16961, 2022.

\bibitem{fort2020DeepEnsembles}
Stanislav Fort, Huiyi Hu, and Balaji Lakshminarayanan.
\newblock Deep {{Ensembles}}: {{A Loss Landscape Perspective}}.
\newblock {\em arXiv:1912.02757}, 2020.

\bibitem{gal2016uncertainty}
Yarin Gal.
\newblock Uncertainty in deep learning.
\newblock {\em Ph.D. thesis, University of Cambridge}, 2016.

\bibitem{gal2016DropoutBayesian}
Yarin Gal and Zoubin Ghahramani.
\newblock Dropout as a bayesian approximation: Representing model uncertainty in deep learning.
\newblock In Maria~Florina Balcan and Kilian~Q. Weinberger, editors, {\em Proceedings of The 33rd International Conference on Machine Learning}, volume~48 of {\em Proceedings of Machine Learning Research}, pages 1050--1059, New York, New York, USA, 20--22 Jun 2016. PMLR.

\bibitem{gal2017deep}
Yarin Gal, Riashat Islam, and Zoubin Ghahramani.
\newblock Deep bayesian active learning with image data.
\newblock In {\em International conference on machine learning}, pages 1183--1192. PMLR, 2017.

\bibitem{gao2022ci}
Tianxiao Gao, Wu Wei, Zhongbin Cai, Zhun Fan, Sheng~Quan Xie, Xinmei Wang, and Qiuda Yu.
\newblock Ci-net: A joint depth estimation and semantic segmentation network using contextual information.
\newblock {\em Applied Intelligence}, 52(15):18167--18186, 2022.

\bibitem{gao2022predictive}
Tianxiao Gao, Wu Wei, Xinmei Wang, Qiuda Yu, and Zhun Fan.
\newblock Predictive uncertainties for multi-task learning network.
\newblock In {\em International Conference on Advanced Algorithms and Neural Networks (AANN 2022)}, volume 12285, pages 294--300. SPIE, 2022.

\bibitem{gawlikowski2022SurveyUncertainty}
Jakob Gawlikowski, Cedrique Rovile~Njieutcheu Tassi, Mohsin Ali, Jongseok Lee, Matthias Humt, Jianxiang Feng, Anna Kruspe, Rudolph Triebel, Peter Jung, Ribana Roscher, Muhammad Shahzad, Wen Yang, Richard Bamler, and Xiao~Xiang Zhu.
\newblock A {{Survey}} of {{Uncertainty}} in {{Deep Neural Networks}}.
\newblock {\em arXiv:2107.03342}, 2022.

\bibitem{guo2017CalibrationModerna}
Chuan Guo, Geoff Pleiss, Yu Sun, and Kilian~Q. Weinberger.
\newblock On calibration of modern neural networks.
\newblock In Doina Precup and Yee~Whye Teh, editors, {\em Proceedings of the 34th International Conference on Machine Learning}, volume~70 of {\em Proceedings of Machine Learning Research}, pages 1321--1330. PMLR, 06--11 Aug 2017.

\bibitem{gurau2018dropout}
Corina Gurau, Alex Bewley, and Ingmar Posner.
\newblock Dropout distillation for efficiently estimating model confidence.
\newblock {\em arXiv preprint arXiv:1809.10562}, 2018.

\bibitem{gustafsson2020evaluating}
Fredrik~K Gustafsson, Martin Danelljan, and Thomas~B Schon.
\newblock Evaluating scalable bayesian deep learning methods for robust computer vision.
\newblock In {\em Proceedings of the IEEE/CVF conference on computer vision and pattern recognition workshops}, pages 318--319, 2020.

\bibitem{he2021sosd}
Lei He, Jiwen Lu, Guanghui Wang, Shiyu Song, and Jie Zhou.
\newblock Sosd-net: Joint semantic object segmentation and depth estimation from monocular images.
\newblock {\em Neurocomputing}, 440:251--263, 2021.

\bibitem{heizmann2022implementing}
Michael Heizmann, Alexander Braun, Markus Glitzner, Matthias G{\"u}nther, G{\"u}nther Hasna, Christina Kl{\"u}ver, Jakob Kroo{\ss}, Erik Marquardt, Michael Overdick, and Markus Ulrich.
\newblock Implementing machine learning: chances and challenges.
\newblock {\em at-Automatisierungstechnik}, 70(1):90--101, 2022.

\bibitem{hinton2015DistillingKnowledgea}
Geoffrey Hinton, Oriol Vinyals, and Jeffrey Dean.
\newblock Distilling the knowledge in a neural network.
\newblock In {\em NIPS Deep Learning and Representation Learning Workshop}, 2015.

\bibitem{Holder_2021_ICCV}
Christopher~J. Holder and Muhammad Shafique.
\newblock Efficient uncertainty estimation in semantic segmentation via distillation.
\newblock In {\em Proceedings of the IEEE/CVF International Conference on Computer Vision (ICCV) Workshops}, pages 3087--3094, October 2021.

\bibitem{hu2018rgb}
Yaosi Hu, Zhenzhong Chen, and Weiyao Lin.
\newblock Rgb-d semantic segmentation: a review.
\newblock In {\em 2018 IEEE International Conference on Multimedia \& Expo Workshops (ICMEW)}, pages 1--6. IEEE, 2018.

\bibitem{huber1992robust}
Peter~J Huber.
\newblock Robust estimation of a location parameter.
\newblock In {\em Breakthroughs in statistics: Methodology and distribution}, pages 492--518. Springer, 1992.

\bibitem{ji2023semantic}
Naihua Ji, Huiqian Dong, Fanyun Meng, and Liping Pang.
\newblock Semantic segmentation and depth estimation based on residual attention mechanism.
\newblock {\em Sensors}, 23(17):7466, 2023.

\bibitem{jiao2018look}
Jianbo Jiao, Ying Cao, Yibing Song, and Rynson Lau.
\newblock Look deeper into depth: Monocular depth estimation with semantic booster and attention-driven loss.
\newblock In {\em Proceedings of the European conference on computer vision (ECCV)}, pages 53--69, 2018.

\bibitem{kendall2017CVUncertainties}
Alex Kendall and Yarin Gal.
\newblock What uncertainties do we need in bayesian deep learning for computer vision?
\newblock In {\em Proceedings of the 31st International Conference on Neural Information Processing Systems}, NIPS'17, page 5580–5590. Curran Associates Inc., 2017.

\bibitem{kendall2018multi}
Alex Kendall, Yarin Gal, and Roberto Cipolla.
\newblock Multi-task learning using uncertainty to weigh losses for scene geometry and semantics.
\newblock In {\em Proceedings of the IEEE conference on computer vision and pattern recognition}, pages 7482--7491, 2018.

\bibitem{lakshminarayanan2017SimpleScalable}
Balaji Lakshminarayanan, Alexander Pritzel, and Charles Blundell.
\newblock Simple and scalable predictive uncertainty estimation using deep ensembles.
\newblock In I. Guyon, U.~Von Luxburg, S. Bengio, H. Wallach, R. Fergus, S. Vishwanathan, and R. Garnett, editors, {\em Advances in Neural Information Processing Systems}, volume~30. {Curran Associates, Inc.}, 2017.

\bibitem{landgraf2023segmentation}
Steven Landgraf, Markus Hillemann, Moritz Aberle, Valentin Jung, and Markus Ulrich.
\newblock Segmentation of industrial burner flames: A comparative study from traditional image processing to machine and deep learning.
\newblock {\em arXiv preprint arXiv:2306.14789}, 2023.

\bibitem{landgraf2023u}
Steven Landgraf, Markus Hillemann, Kira Wursthorn, and Markus Ulrich.
\newblock U-ce: Uncertainty-aware cross-entropy for semantic segmentation.
\newblock {\em arXiv preprint arXiv:2307.09947}, 2023.

\bibitem{landgraf2023dudes}
Steven Landgraf, Kira Wursthorn, Markus Hillemann, and Markus Ulrich.
\newblock Dudes: Deep uncertainty distillation using ensembles for semantic segmentation.
\newblock {\em arXiv preprint arXiv:2303.09843}, 2023.

\bibitem{lee2018TrainingConfidencecalibrated}
Kimin Lee, Honglak Lee, Kibok Lee, and Jinwoo Shin.
\newblock Training {{Confidence-calibrated Classifiers}} for {{Detecting Out-of-Distribution Samples}}.
\newblock {\em arXiv:1711.09325}, 2018.

\bibitem{leibig2017LeveragingUncertainty}
Christian Leibig, Vaneeda Allken, Murat~Se{\c c}kin Ayhan, Philipp Berens, and Siegfried Wahl.
\newblock Leveraging uncertainty information from deep neural networks for disease detection.
\newblock {\em Scientific Reports}, 7(1):17816, 2017.

\bibitem{lin2019depth}
Xiao Lin, Dalila S{\'a}nchez-Escobedo, Josep~R Casas, and Montse Pard{\`a}s.
\newblock Depth estimation and semantic segmentation from a single rgb image using a hybrid convolutional neural network.
\newblock {\em Sensors}, 19(8):1795, 2019.

\bibitem{liu2020simple}
Jeremiah Liu, Zi Lin, Shreyas Padhy, Dustin Tran, Tania Bedrax~Weiss, and Balaji Lakshminarayanan.
\newblock Simple and principled uncertainty estimation with deterministic deep learning via distance awareness.
\newblock {\em Advances in Neural Information Processing Systems}, 33:7498--7512, 2020.

\bibitem{liu2018collaborative}
Jing Liu, Yuhang Wang, Yong Li, Jun Fu, Jiangyun Li, and Hanqing Lu.
\newblock Collaborative deconvolutional neural networks for joint depth estimation and semantic segmentation.
\newblock {\em IEEE transactions on neural networks and learning systems}, 29(11):5655--5666, 2018.

\bibitem{liu2019end}
Shikun Liu, Edward Johns, and Andrew~J Davison.
\newblock End-to-end multi-task learning with attention.
\newblock In {\em Proceedings of the IEEE/CVF conference on computer vision and pattern recognition}, pages 1871--1880, 2019.

\bibitem{loquercio2020general}
Antonio Loquercio, Mattia Segu, and Davide Scaramuzza.
\newblock A general framework for uncertainty estimation in deep learning.
\newblock {\em IEEE Robotics and Automation Letters}, 5(2):3153--3160, 2020.

\bibitem{loshchilov2017decoupled}
Ilya Loshchilov and Frank Hutter.
\newblock Decoupled weight decay regularization.
\newblock {\em arXiv preprint arXiv:1711.05101}, 2017.

\bibitem{mackay1992PracticalBayesian}
David J.~C. MacKay.
\newblock A {{Practical Bayesian Framework}} for {{Backpropagation Networks}}.
\newblock {\em Neural Computation}, 4(3):448--472, 1992.

\bibitem{malinin2019EnsembleDistributiona}
Andrey Malinin, Bruno Mlodozeniec, and Mark Gales.
\newblock Ensemble {{Distribution Distillation}}.
\newblock {\em arXiv:1905.00076}, 2019.

\bibitem{mcallister2017ConcreteProblems}
Rowan McAllister, Yarin Gal, Alex Kendall, Mark {van der Wilk}, Amar Shah, Roberto Cipolla, and Adrian Weller.
\newblock Concrete {{Problems}} for {{Autonomous Vehicle Safety}}: {{Advantages}} of {{Bayesian Deep Learning}}.
\newblock In {\em Proceedings of the {{Twenty-Sixth International Joint Conference}} on {{Artificial Intelligence}}}, pages 4745--4753, {Melbourne, Australia}, 2017. {International Joint Conferences on Artificial Intelligence Organization}.

\bibitem{micikevicius2017mixed}
Paulius Micikevicius, Sharan Narang, Jonah Alben, Gregory Diamos, Erich Elsen, David Garcia, Boris Ginsburg, Michael Houston, Oleksii Kuchaiev, Ganesh Venkatesh, et~al.
\newblock Mixed precision training.
\newblock {\em arXiv preprint arXiv:1710.03740}, 2017.

\bibitem{minaee2020ImageSegmentation}
Shervin Minaee, Yuri Boykov, Fatih Porikli, Antonio Plaza, Nasser Kehtarnavaz, and Demetri Terzopoulos.
\newblock Image segmentation using deep learning: A survey.
\newblock {\em IEEE Transactions on Pattern Analysis and Machine Intelligence}, 44(7):3523--3542, 2022.

\bibitem{ming2021deep}
Yue Ming, Xuyang Meng, Chunxiao Fan, and Hui Yu.
\newblock Deep learning for monocular depth estimation: A review.
\newblock {\em Neurocomputing}, 438:14--33, 2021.

\bibitem{mousavian2016joint}
Arsalan Mousavian, Hamed Pirsiavash, and Jana Ko{\v{s}}eck{\'a}.
\newblock Joint semantic segmentation and depth estimation with deep convolutional networks.
\newblock In {\em 2016 Fourth International Conference on 3D Vision (3DV)}, pages 611--619. IEEE, 2016.

\bibitem{mukhoti2018evaluating}
Jishnu Mukhoti and Yarin Gal.
\newblock Evaluating bayesian deep learning methods for semantic segmentation.
\newblock {\em arXiv preprint arXiv:1811.12709}, 2018.

\bibitem{mukhoti2023deep}
Jishnu Mukhoti, Andreas Kirsch, Joost van Amersfoort, Philip~HS Torr, and Yarin Gal.
\newblock Deep deterministic uncertainty: A new simple baseline.
\newblock In {\em Proceedings of the IEEE/CVF Conference on Computer Vision and Pattern Recognition}, pages 24384--24394, 2023.

\bibitem{naeini2015obtaining}
Mahdi~Pakdaman Naeini, Gregory Cooper, and Milos Hauskrecht.
\newblock Obtaining well calibrated probabilities using bayesian binning.
\newblock In {\em Proceedings of the AAAI conference on artificial intelligence}, volume~29, 2015.

\bibitem{nekrasov2019real}
Vladimir Nekrasov, Thanuja Dharmasiri, Andrew Spek, Tom Drummond, Chunhua Shen, and Ian Reid.
\newblock Real-time joint semantic segmentation and depth estimation using asymmetric annotations.
\newblock In {\em 2019 International Conference on Robotics and Automation (ICRA)}, pages 7101--7107. IEEE, 2019.

\bibitem{nix1994estimating}
David~A Nix and Andreas~S Weigend.
\newblock Estimating the mean and variance of the target probability distribution.
\newblock In {\em Proceedings of 1994 ieee international conference on neural networks (ICNN'94)}, volume~1, pages 55--60. IEEE, 1994.

\bibitem{ovadia2019DatasetShift}
Yaniv Ovadia, Emily Fertig, Jie Ren, Zachary Nado, D. Sculley, Sebastian Nowozin, Joshua Dillon, Balaji Lakshminarayanan, and Jasper Snoek.
\newblock Can you trust your model\textquotesingle s uncertainty? evaluating predictive uncertainty under dataset shift.
\newblock In H. Wallach, H. Larochelle, A. Beygelzimer, F. d\textquotesingle Alch\'{e}-Buc, E. Fox, and R. Garnett, editors, {\em Advances in Neural Information Processing Systems}, volume~32. Curran Associates, Inc., 2019.

\bibitem{romero2015FitNetsHints}
Adriana Romero, Nicolas Ballas, Samira~Ebrahimi Kahou, Antoine Chassang, Carlo Gatta, and Yoshua Bengio.
\newblock {{FitNets}}: {{Hints}} for {{Thin Deep Nets}}.
\newblock {\em arXiv:1412.6550}, 2015.

\bibitem{schwaiger2020uncertainty}
Adrian Schwaiger, Poulami Sinhamahapatra, Jens Gansloser, and Karsten Roscher.
\newblock Is uncertainty quantification in deep learning sufficient for out-of-distribution detection?
\newblock {\em Aisafety@ ijcai}, 54, 2020.

\bibitem{Shen_2021_WACV}
Yichen Shen, Zhilu Zhang, Mert~R. Sabuncu, and Lin Sun.
\newblock Real-time uncertainty estimation in computer vision via uncertainty-aware distribution distillation.
\newblock In {\em Proceedings of the IEEE/CVF Winter Conference on Applications of Computer Vision (WACV)}, pages 707--716, January 2021.

\bibitem{shrivastava2016training}
Abhinav Shrivastava, Abhinav Gupta, and Ross Girshick.
\newblock Training region-based object detectors with online hard example mining.
\newblock In {\em Proceedings of the IEEE conference on computer vision and pattern recognition}, pages 761--769, 2016.

\bibitem{silberman2012indoor}
Nathan Silberman, Derek Hoiem, Pushmeet Kohli, and Rob Fergus.
\newblock Indoor segmentation and support inference from rgbd images.
\newblock In {\em Computer Vision--ECCV 2012: 12th European Conference on Computer Vision, Florence, Italy, October 7-13, 2012, Proceedings, Part V 12}, pages 746--760. Springer, 2012.

\bibitem{simpson2022learning}
Ivor~JA Simpson, Sara Vicente, and Neill~DF Campbell.
\newblock Learning structured gaussians to approximate deep ensembles.
\newblock In {\em Proceedings of the IEEE/CVF Conference on Computer Vision and Pattern Recognition}, pages 366--374, 2022.

\bibitem{srivastava2014Dropout}
Nitish Srivastava, Geoffrey Hinton, Alex Krizhevsky, Ilya Sutskever, and Ruslan Salakhutdinov.
\newblock Dropout: A simple way to prevent neural networks from overfitting.
\newblock {\em Journal of Machine Learning Research}, 15(56):1929--1958, 2014.

\bibitem{steger2018MachineVision}
Carsten Steger, Markus Ulrich, and Christian Wiedemann.
\newblock {\em Machine Vision Algorithms and Applications}.
\newblock John Wiley \& Sons, 2018.

\bibitem{valdenegro2023sub}
Matias Valdenegro-Toro.
\newblock Sub-ensembles for fast uncertainty estimation in neural networks.
\newblock In {\em Proceedings of the IEEE/CVF International Conference on Computer Vision}, pages 4119--4127, 2023.

\bibitem{van2020uncertainty}
Joost Van~Amersfoort, Lewis Smith, Yee~Whye Teh, and Yarin Gal.
\newblock Uncertainty estimation using a single deep deterministic neural network.
\newblock In {\em International conference on machine learning}, pages 9690--9700. PMLR, 2020.

\bibitem{vandenhende2020mti}
Simon Vandenhende, Stamatios Georgoulis, and Luc Van~Gool.
\newblock Mti-net: Multi-scale task interaction networks for multi-task learning.
\newblock In {\em Computer Vision--ECCV 2020: 16th European Conference, Glasgow, UK, August 23--28, 2020, Proceedings, Part IV 16}, pages 527--543. Springer, 2020.

\bibitem{wang2021brief}
Changshuo Wang, Chen Wang, Weijun Li, and Haining Wang.
\newblock A brief survey on rgb-d semantic segmentation using deep learning.
\newblock {\em Displays}, 70:102080, 2021.

\bibitem{wang2015towards}
Peng Wang, Xiaohui Shen, Zhe Lin, Scott Cohen, Brian Price, and Alan~L Yuille.
\newblock Towards unified depth and semantic prediction from a single image.
\newblock In {\em Proceedings of the IEEE conference on computer vision and pattern recognition}, pages 2800--2809, 2015.

\bibitem{wursthorn2022}
Kira Wursthorn, Markus Hillemann, and Markus Ulrich.
\newblock Comparison of uncertainty quantification methods for {CNN}-based regression.
\newblock {\em The International Archives of the Photogrammetry, Remote Sensing and Spatial Information Sciences}, XLIII-B2-2022:721--728, 2022.

\bibitem{xie2021segformer}
Enze Xie, Wenhai Wang, Zhiding Yu, Anima Anandkumar, Jose~M Alvarez, and Ping Luo.
\newblock Segformer: Simple and efficient design for semantic segmentation with transformers.
\newblock {\em Advances in Neural Information Processing Systems}, 34:12077--12090, 2021.

\bibitem{xu2018pad}
Dan Xu, Wanli Ouyang, Xiaogang Wang, and Nicu Sebe.
\newblock Pad-net: Multi-tasks guided prediction-and-distillation network for simultaneous depth estimation and scene parsing.
\newblock In {\em Proceedings of the IEEE Conference on Computer Vision and Pattern Recognition}, pages 675--684, 2018.

\bibitem{xu2022mtformer}
Xiaogang Xu, Hengshuang Zhao, Vibhav Vineet, Ser-Nam Lim, and Antonio Torralba.
\newblock Mtformer: Multi-task learning via transformer and cross-task reasoning.
\newblock In {\em European Conference on Computer Vision}, pages 304--321. Springer, 2022.

\bibitem{zhang2021survey}
Yu Zhang and Qiang Yang.
\newblock A survey on multi-task learning.
\newblock {\em IEEE Transactions on Knowledge and Data Engineering}, 34(12):5586--5609, 2021.

\end{thebibliography}
}

\end{document}